# DESIGN AND IMPLEMENTATION
# OF
# A TACTICAL GENERATOR
# FOR TURKISH,
# A FREE CONSTITUENT ORDER LANGUAGE

A THESIS
SUBMITTED TO THE DEPARTMENT OF COMPUTER ENGINEERING
AND INFORMATION SCIENCE
AND THE INSTITUTE OF ENGINEERING AND SCIENCE
OF BILKENT UNIVERSITY
IN PARTIAL FULFILLMENT OF THE REQUIREMENTS
FOR THE DEGREE OF
MASTER OF SCIENCE

By
Dilek Zeynep Hakkani
July, 1996

I certify that I have read this thesis and that in my opinion it is fully adequate, in scope and in quality, as a thesis for the degree of Master of Science.

Dr. Kemal Oflazer (Principal Advisor)

I certify that I have read this thesis and that in my opinion it is fully adequate, in scope and in quality, as a thesis for the degree of Master of Science.

Dr. İlyas Çiçekli (Co-advisor)

I certify that I have read this thesis and that in my opinion it is fully adequate, in scope and in quality, as a thesis for the degree of Master of Science.

Dr. Halil Altay Güvenir

Approved for the Institute of Engineering and Science:

Prof. Dr. Mehmet Baray, Director of Institute of Engineering and Science

# ABSTRACT

**DESIGN AND IMPLEMENTATION
OF
A TACTICAL GENERATOR
FOR TURKISH,
A FREE CONSTITUENT ORDER LANGUAGE**

Dilek Zeynep Hakkani

M.S. in Computer Engineering and Information Science

Principal Advisor: Asst. Prof. Kemal Oflazer

Co-advisor: Asst. Prof. İlyas Çiçekli

July, 1996

This thesis describes a tactical generator for Turkish, a free constituent order language, in which the order of the constituents may change according to the information structure of the sentences to be generated. In the absence of any information regarding the information structure of a sentence (i.e., topic, focus, background, etc.), the constituents of the sentence obey a default order, but the order is almost freely changeable, depending on the constraints of the text flow or discourse. We have used a recursively structured finite state machine for handling the changes in constituent order, implemented as a right-linear grammar backbone. Our implementation environment is the GenKit system, developed at Carnegie Mellon University–Center for Machine Translation. Morphological realization has been implemented using an external morphological analysis/generation component which performs concrete morpheme selection and handles morphographemic processes.

# ÖZET

## SERBEST ÖĞE SIRALI BİR DİL OLAN TÜRKÇE İÇİN YÜZEYSEL ÜRETİCİ TASARIMI VE GERÇEKLEŞTİRMESİ

Dilek Zeynep Hakkani

Bilgisayar ve Enformatik Mühendisliği, Yüksek Lisans

Tez Yöneticisi: Yrd. Doç. Dr. Kemal Oflazer

Yardımcı Tez Yöneticisi: Yrd. Doç. Dr. İlyas Çiçekli

Temmuz, 1996

Bu tezde, gerçekleştirimi serbest öğe düzenine sahip bir dil olan Türkçe için bir yüzeysel üretici sunulmaktadır. Bir cümlenin bilgi yapısıyla ilgili herhangi bir bilginin (başlık, odak, arka plan, v.b.g.) olmaması durumunda, tümce öğeleri öngörülen bir sıraya uyarlar. Ancak bu sıra, tümcenin akışı veya konuşmanın sınırlamalarına göre serbetçe değişebilir. Öğelerin sırasındaki değişiklikleri üretmek için sağ doğrusal gramer ile gerçekleştirilmiş bir öz yinelemeli sonlu durum makinesi kullanılmıştır. Gerçekleştirme ortamımız, Carnegie Mellon Üniversitesi – Center for Machine Translation'da (CMU - CMT) geliştirilen GenKit sistemidir. Biçimbirimsel gerçekleştirme, dışarıdan çağırılan, somut biçimbirim seçimini ve biçimbirimsel değişmeleri sağlayan biçimbirimsel bir üretim sistemi kullanılarak gerçekleştirilmiştir.

*Anahtar sözcükler*: Doğal Dil Üretimi, Serbest Öğe Sıralı Diller, Gerçekleştirme, Gramer Teorisi.

To my family

# ACKNOWLEDGEMENTS

I am very grateful to my co-advisors, Assistant Professor Kemal Oflazer and Assistant Professor İlyas Çiçekli, who have provided invaluable guidance during this study.

I would like to thank to Assistant Professor Kemal Oflazer for providing me a stimulating research environment. His instruction will be the closest and most important reference in my future research.

I would also like to thank Assoc. Prof. Halil Altay Güvenir for his valuable comments, and guidance on this thesis.

I would like to thank Dr. Beryl Hoffman of University of Edinburgh, Centre for Cognitive Science, for her kind comments on certain aspects and presentation of this work. This research has been supported in part by a NATO Science for Stability Project Grant TU–LANGUAGE.

I would like to thank to my family. I am very grateful for their moral support and hope-giving. I would like to thank everybody who has in some way contributed to this study by lending me moral, technical and intellectual support, including my sister Selin Hakkani and my colleagues Bilge Say, Yücel Saygın, Kemal Ülkü, A. Kurtuluş Yorulmaz, Gökhan Tür and many others who are not mentioned here by name.

Finally, I would like to thank to Mr. Gökhan Tür again, for always being with me.

# Contents

# List of Figures

# List of Abbreviations

| | |
|---|---|
| 1SG, 2SG, 3SG | first, second, third person singular |
| 1PL, 2PL, 3PL | first, second, third person plural |
| P1SG, P2SG, P3SG | first, second, third person singular possessive |
| P1PL, P2PL, P3PL | first, second, third person plural possessive |
| ABILITY | positive potential ($+yAbIl$) |
| ABL | ablative ($+dAyn$) |
| ACC | accusative ($+yH$) |
| ADVB | adverbial conversion |
| AOR | aorist (positive: $+Ar$ and $+Hr$; negative: $+z$) |
| COPULA | copula ($+dIr$) |
| COND | conditional ($+sA$) |
| DAT | dative ($+yA$) |
| FUT | future ($+yAcAk$) |
| GEN | genitive ($+nHn$) |
| INF | infinitive ($+mAk$) |
| LOC | locative ($+dA$) |
| NEG | verbal negative ($+mA$) |
| PART | participle conversion |
| PAST | past ($+dH$) |
| PLU | noun plural ($+lAr$) |
| PRG | progressive ($+Iyor$) |
| QUES | yes/no question ($mH$) |
| REL | relativization ($+ki$) |

# Chapter 1

# Introduction

Natural Language Processing (NLP) is a research area which investigates computational systems that analyze, understand, process, and produce natural language. Every NLP system has one or both of the following subsystems:

- **Parser:** a component which analyzes natural language sentences, and converts them into representations that can further be processed by the computer.

- **Generator:** a component which produces natural language sentences from computer internal representations.

Some applications of NLP systems are: machine translation systems, interfaces to database systems, speech understanding and production systems, and text skimming systems. In a machine translation system, the computer analyzes a given text in one language (called the source language), and then produces the translation of this text in another language (called the target language). The production of text in the target language is done by a natural language generation (NLG) system. In transfer-based machine translation, the generation system is a tactical generator. NLG systems also produce the results of the queries in a natural language in NLP interfaces to database, and the summaries of analyzed text in text skimming systems.

As a component of a large-scale project on natural language processing for Turkish, we have undertaken the development of a tactical generator. This tactical generator can be used in machine translation applications, or in other natural language generation systems together with a strategic generator.

We currently plan to use this tactical generator in prototype transfer-based human-assisted machine translation system from English to Turkish. The analysis of the English sentences is done by an English analysis component. The transfer component transfers the output of the English analyzer, a representation for an English sentence, into a representation for a Turkish sentence. The tactical generator, then, generates the surface form of the Turkish sentence, which is the translation of the input sentence into Turkish. An outline of this system is given in Figure 1.1. For example, if the English sentence "The man wanted to read the book" is given as input to the English analysis component, it produces the following case-frame representation:[1]

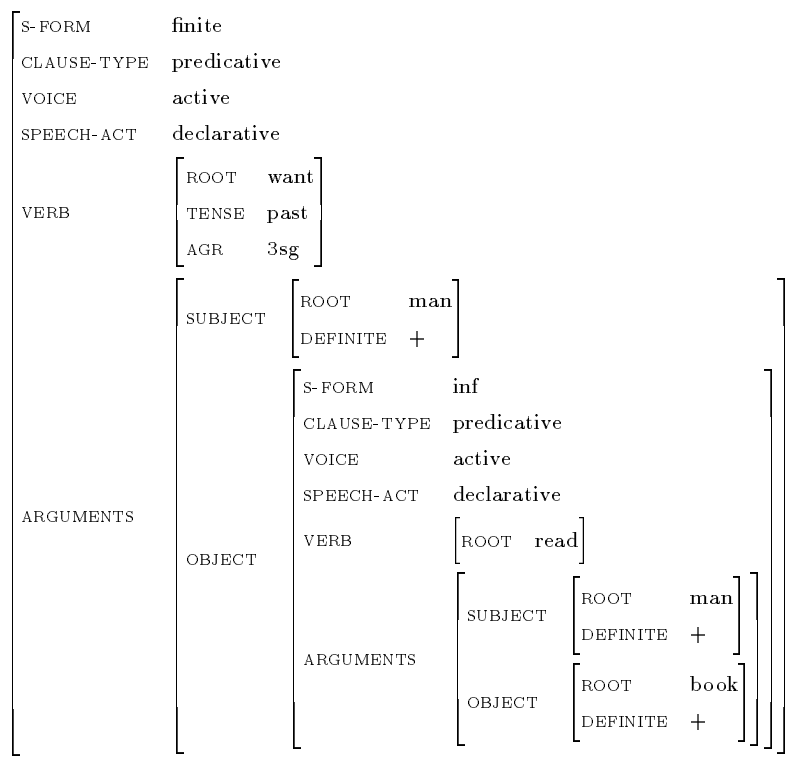

---

[1] We use the case-frame representation as a computer internal representation. A case-frame is a common representation for capturing the predication, arguments, and adjuncts involved in a sentence. We give the details of our case-frames in Chapter 4.

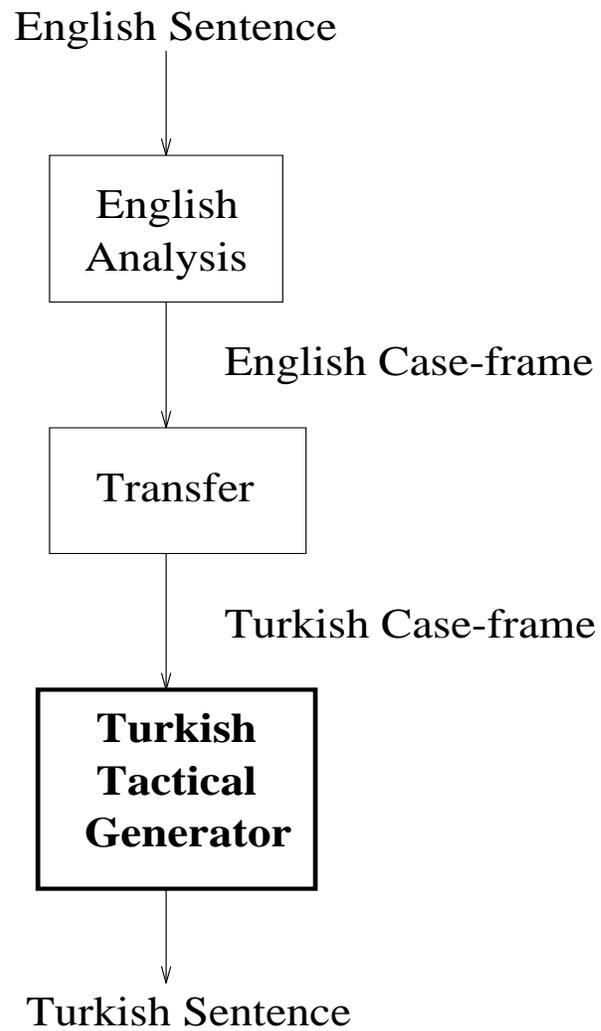

Figure 1.1: The outline of the machine translation project.

As can be seen above, this case-frame has the features, VERB, to capture the predication, and ARGUMENTS, to capture the arguments. The object of the sentence above is also a sentential clause, so the value of the OBJECT feature is a structure very similar to the case-frame.

The transfer component converts this case-frame into another case-frame representing a Turkish sentence, which is the following case-frame:

```
((s-form finite)
 (clause-type predicative)
 (voice active)
 (speech-act declarative)
 (verb
  ((root "iste")
   (sense positive)
   (tense past)
   (aspect perfect)))
 (arguments
  ((subject
     ((referent
        ((arg
           ((concept "adam")))
         (agr
          ((number singular)
           (person 3))))))))
   (dir-obj
    ((roles
      ((role act)
       (arg
        ((s-form inf-ind-act)
         (clause-type predicative)
         (voice active)
         (speech-act declarative)
         (verb
```

```
           ((root "oku")
            (sense positive)))
         (arguments
          ((dir-obj
             ((referent
                ((arg
                   ((concept "kitap")))
                 (agr
                   ((number singular)
                    (person 3)))))
              (specifier
                ((quan
                   ((definite +)))))))))))
```

Then the tactical generator generates the Turkish sentence "Adam kitabı okumak istedi." from this case-frame. During this generation process, it uses a Turkish grammar and lexicon, and imposes the right word order and generates the relevant morphological features.

Turkish, our target language, can be considered as a *subject-object-verb* (SOV) language, in which constituents can change order rather freely, at certain phrase levels, depending on the constraints of text flow or discourse. This constituent order freeness comes from the fact that the morphology of Turkish enables morphological markings on the constituents to express their grammatical roles without relying on their order.

To develop a tactical generator for Turkish, we have used a recursively structured finite state machine, which handles constituent order changes. As the surface constituent order is almost freely changeable depending on the constraints of the text flow or discourse, these constraints obtained from the information structures in the case-frames of the sentences to be generated guide the generator to emit the proper word order. In the absence of any information structure, the constituents of the sentence obey a default order.

Our implementation environment is the GenKit generation system [22], developed at Carnegie Mellon University–Center for Machine Translation. Morphological realization has been implemented using an external morphological analysis/generation component which performs concrete morpheme selection and handles morphographemic processes.

## 1.1 Overview of the Thesis

The outline of the thesis is as follows: Chapter 2 describes briefly Natural Language Generation, and its phases (text planning, sentence planning, and realization), and the scope of our work. Chapter 3 presents an overview of Turkish syntax, emphasizing the concepts that we dealt with when designing the grammar. Chapter 4 describes our approach for generating Turkish sentences, together with the architecture of our grammar. We also provide here a comparison of our work with related work on Turkish. Chapter 5 presents an evaluation of our grammar with some example outputs of the generator, along with proposals for future work. Chapter 6 concludes the thesis.

# Chapter 2

# Natural Language Generation

Natural language generation is the process of producing natural language sentences using specified communicative goals [15]. This area of study investigates the way computer programs can produce high-quality natural language text from computer-internal representations of information [12]. The natural language generation process consists of three main activities [18]:

1. The information that should be communicated to the user and the way this information should be structured must be determined. These, usually simultaneous, tasks are called as *content determination* and *text planning*, respectively.

2. The split of information among individual sentences and paragraphs must be determined (sentence planning). During this process, in order to make a smooth text flow, *cohesion devices* (such as pronouns) to be added, should be dictated.

3. The individual sentences should be generated in a grammatically correct manner (realization).

In most natural language generation systems there are two different parts [2, 25]:

1. the *strategic generator*, which implements the first two of the activities above, and

2. the *tactical generator*, which implements the last one of the activities above.

In the remaining of this chapter, we present an overview of strategic generation and tactical generation, followed by a description of the scope of our work.

## 2.1 Strategic Generation

As indicated above, the first two activities in natural language generation, that of identifying the goals the utterance is to achieve and planning the way these goals may be achieved, is called as strategic generation [15]. For example, in order to describe the event in the picture in Figure 2.1, at least these five Turkish sentences can be generated:

```
a) Ali kitabı Ahmet'e verdi.
b) Kitap Ahmet'e verildi.
c) Kitap Ahmet'e Ali tarafından verildi.
d) Ahmet'e kitap verildi.
e) Ali Ahmet'e kitabı verdi.
```

A strategic generator determines which (words or concepts and) surface form you would use to describe this event, taking into account information that is mostly not linguistic, such as:

- world knowledge,
- previous context in discourse,
- speaker intentions, etc.

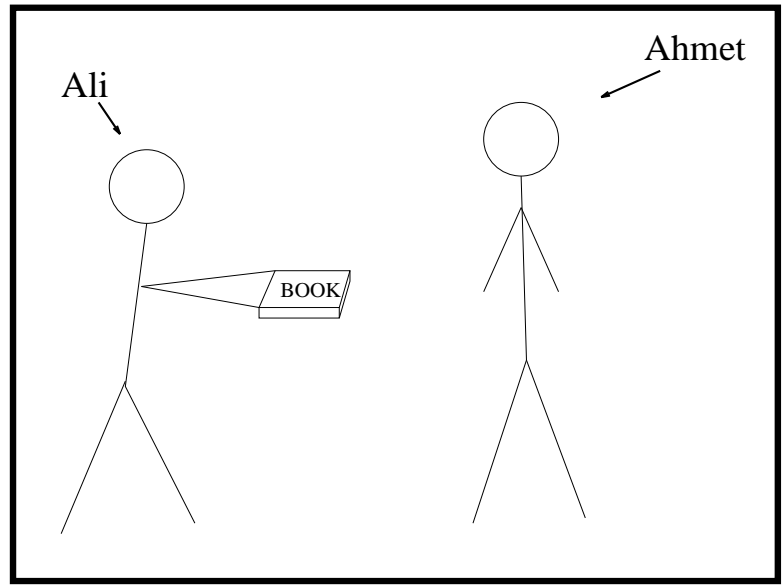

Figure 2.1: A picture for demonstrating the event of Ali's giving the book to Ahmet.

## 2.2 Tactical Generation

The tactical generator, realizes, as linear text, the contents of a sentence which are specified usually using some kind of a feature structure. This feature structure can be generated by a higher level process such as a strategic generator or transfer process in machine translation applications, as demonstrated in Figure 2.2. In this process a generation grammar and a generation lexicon are used.

A natural language grammar is a formal device for defining the relation between natural language utterances and the computer-internal representations to express their meaning [26]. The same grammar (a reversible grammar) can be used for both analysis and generation. But, problems of parsing and generation are rather different. In parsing, ambiguity at all levels (lexical, syntactic, semantic) is a very serious problem. Whereas, in generation, the problem is non-determinism, production of more than one sentence from a computer-internal representation. A generation grammar must also be augmented with style-related information. However, this information can be ignored in an analysis grammar, if only the semantics of the sentence is needed [5].

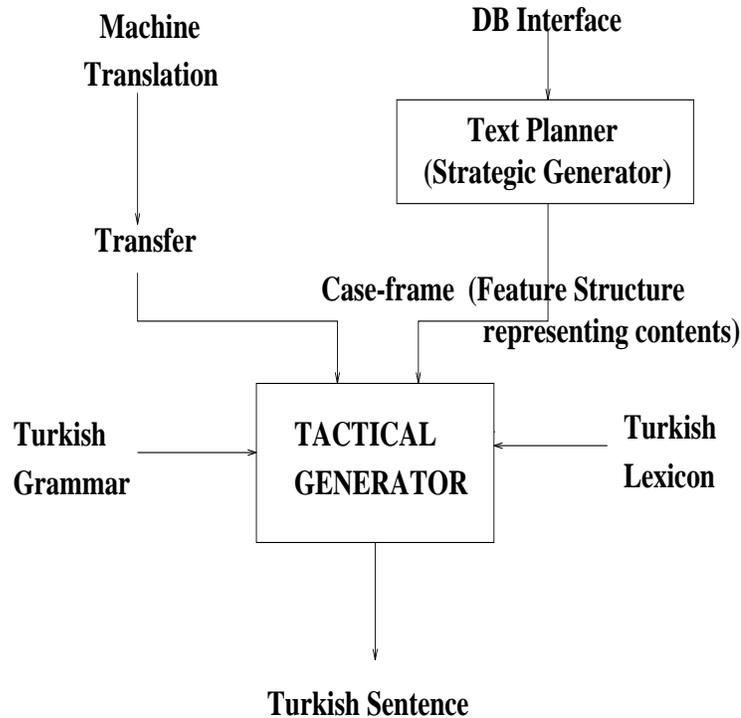

Figure 2.2: The usage of a tactical generator

## 2.3 Scope of Our Work

Our main goal in this thesis is to develop a tactical generator for Turkish that we can use in a prototype machine translation system from English to Turkish. Our tactical generator gets a feature structure as input from the transfer component in this machine translation system, representing the contents of the sentence to be generated, where all lexical choices have been made. The feature structures for these sentences are represented using a case-frame representation that will be detailed later. This information is then converted into a linear sequence of lexical feature structures. Then, in order to perform morphological realization, this output of the tactical generator is sent to an external morphological generation component which performs concrete morpheme selection and handles morphographemic phenomena such as vowel harmony, and vowel and consonant ellipsis and then produces an agglutinative surface form. As Turkish morphology is outside the scope of this work, we refer the reader to relevant work [17]. The interface of our tactical generator with other components is shown in Figure 2.3.

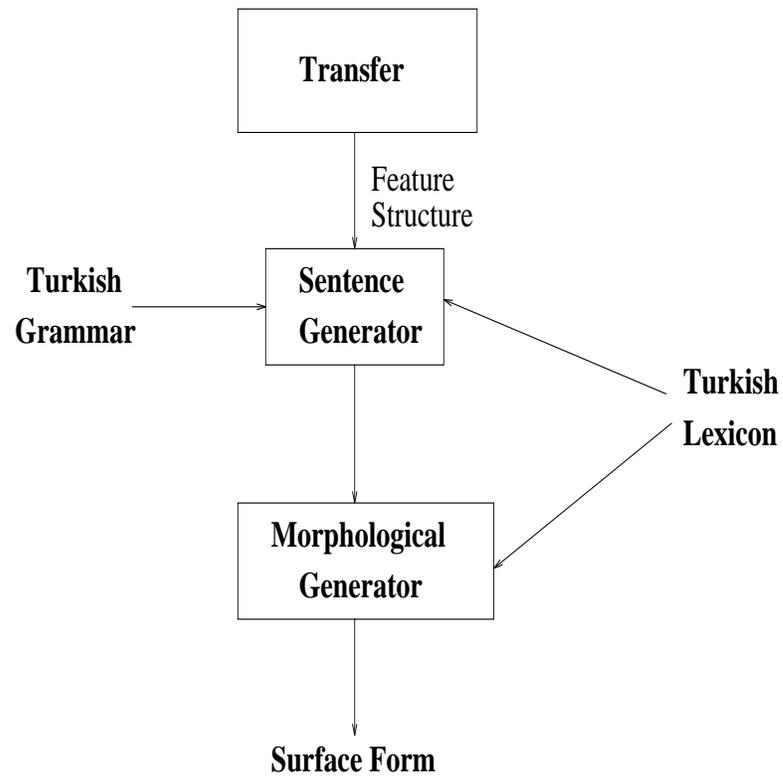

Figure 2.3: The interface of our tactical generator

# Chapter 3

# Turkish

Turkish is a free constituent order language, in which the order of the constituents may change according to the information to be conveyed. In the absence of any information regarding the information structure of a sentence (i.e., topic, focus, background, etc.), the constituents of the sentence obey a default order, but otherwise the order is almost freely changeable, depending on the constraints of the text flow or discourse. In the next section, we present the components of the information structure which controls the constituent order variations and an overview of Turkish sentences, noun phrases, and sentential clauses, relevant to subsequent chapters.

## 3.1 Information Structure

In terms of word order, Turkish can be characterized as a *subject–object–verb (SOV) language* in which constituents at certain phrase levels can change order rather freely, depending on the constraints of text flow or discourse. This is due to the fact that the morphology of Turkish enables morphological markings on the constituents to signal their grammatical roles without relying on their order. For example, the word 'kitap' (book) case marked accusative is a definite direct object, the word 'ev' (house) case marked dative expresses a goal and the word 'okul' (school) case marked ablative expresses a source:

    kitab+ı              (Definite dir. object – theme)
    `book+ACC`
    ev+e                (Dative object – goal)
    `house+DAT`
    okul+dan          (Ablative object – source)
    `school+ABL`

This, however, does not mean that word order is immaterial. Sentences with different word orders reflect different pragmatic conditions, in that, topic, focus and background information conveyed by such sentences differ.[1] Information conveyed through intonation, stress and/or clefting in fixed word order languages such as English, is expressed in Turkish by changing the order of the constituents. Obviously, there are certain constraints on constituent order, especially, inside noun and post-positional phrases. There are also certain constraints at sentence level when explicit case marking is not used (e.g., with indefinite direct objects).

Information structure indicates how linguistically conveyed information is to be added to a context (hearer's information state) [24]. In free word order languages, the information structure (topic, focus, and background) is indicated by the word order [8]. In Turkish, the information which links the sentence to the previous context, the *topic*, is in the first position [4]. For example, in the sentence (b) below, the direct object, which is a pronoun, is the topic of that sentence:[2]

(1) a. Ayşe evde    çok   sıkıldı.
      `Ayşe home+LOC very get-bored+PAST+3SG`
      *'Ayşe got very bored at home.'*

---

[1]See Erguvanlı [4] for a discussion of the function of word order in Turkish grammar.

[2]In the glosses, `3SG` denotes third person singular verbal agreement, `P1PL` and `P3SG` denote first person plural and third person singular possessive agreement, `WITH` denotes a derivational marker making adjectives from nouns, `LOC, ABL, DAT, GEN` denote locative, ablative, dative, and genitive case markers, `PAST` denotes past tense, and `INF` denotes a marker that derives an infinitive form from a verb.

    b. Onu     da   sinemaya     çağırabilir miyim?
       `She+ACC too cinema+DAT call+ABILITY+QUES+1SG`
       *'Can I call her to the cinema too?'*

The information which is new or emphasized, the *focus*, is in the immediately preverbal position [4]. For example, in the answer to the following question, the subject, "her mother", is the focus:

(2) Q: Bu    topu       Ayşe'ye    kim aldı?
      `this ball+ACC Ayşe+DAT who buy+PAST+3SG`
      *'Who bought this ball to Ayşe?'*

    A: Bu    topu       Ayşe'ye    annesi       aldı.
       `this ball+ACC Ayşe+DAT mother+P3SG buy+PAST+3SG`
       *'Her mother bought this ball for Ayşe.'*

The additional information which may be given to help the hearer understand the sentence, the *background*, is in the post verbal position [4]. For example, in the second sentence below, the subject, "Ayşe", which is also the subject and the topic of the first sentence, is the background.

(3) a. Ayşe bütün kitaplarını       eve       götürmek
      `Ayşe all   book+PLU+P3SG+ACC home+DAT bring+INF`
      *'Ayşe wanted to bring all her books to home. '*

      istedi.
      `want+PAST+3SG`

    b. Fakat, tarih    kitabını       okulda       unuttu         Ayşe.
      `but   history book+P3SG+ACC school+LOC forget+PAST+3SG Ayşe`
      *'But she, Ayşe, forgot her history book at school.'*

Thus the topic, focus and background information, when available, alter the order of constituents of Turkish sentences.

## 3.2  Simple Sentences

Turkish sentences can be grouped into three:

1. *predicative sentences,*
2. *existential sentences,*
3. *attributive sentences.*

Predicative sentences have predicative verbs inflected in the verb paradigm. The following are some example predicative sentences:

(4) a. Kitaplarımı       sınıfta        unuttum.
       `book+PLU+P1SG+ACC classroom+LOC forget+PAST+1SG`
       *'I forgot my books in the classroom.'*

   b. Çocukları     okula      anneleri    getirdi.
       `child+PLU+ACC school+DAT mother+P3PL bring+PAST+3SG`
       *'Their mother brought the children to the school.'*

Existential sentences have verbs denoting existence ('var' in Turkish) or nonexistence ('yok' in Turkish).[3] The following are some example existential sentences:

(5) a. Benim iki kalemim    var.
       `I+GEN two pencil+P1SG existent`
       *'I have two pencils.'*

   a. Odasında       perde  bile yoktu.
       `room+P3SG+LOC curtain even non-existent+PAST`
       *'There were even no curtains in her room.'*

Attributive sentences have nominal verbs which express some property of the subject noun phrase. The following are some example attributive sentences:

---

[3]These correspond to *'There is/are ...'* sentences in English.

(6) a. Bu　çay　çok　sıcak.
　　　this tea very hot
　　　'This tea is very hot.'

　b. Kitaplar　masamın　　　üzerinde.
　　　book+PLU table+P1SG+GEN on+P3SG+LOC
　　　'The books are on my table.'

In the following sections, we present additional information about these three kinds of sentences including their constituents, default word order, etc.

### 3.2.1　Predicative Sentences

Predicative sentences are sentences whose verbs are inflected in the verb paradigm. *Typical* constituents of such sentences are: subject, expression of time, expression of place, direct object, beneficiary, source, goal, location, instrument, value designator, path, duration, expression of manner and verb (the verb is obligatory).

- The *subject* is the syntactic subject.

- *Expression of time* and *expression of place* are adjuncts.

- *Direct object* is the syntactic direct object of the sentence.

- *Beneficiary* is the person who is benefiting from an action or a state.

- *Source* indicates the point of origin of a displacement, whereas *goal* indicates the destination of a displacement.

- *Location* denotes the spatial position of the predicate.

- *Instrument* is the medium which the predicate is done with.

- *Value designator* is the money which the action is taken for.

- *Path, duration*, and *expression of manner* are adjuncts.

In the absence of any control information, such as the information structure components topic, focus, or background, (discussed earlier, indicating discourse constraints) the constituents of Turkish sentences have the following default order:

> *subject, expression of time, expression of place, direct object, beneficiary, source, goal, location, instrument, value designator, path, duration, expression of manner, verb.*

All of these constituents except the verb are optional *unless the verb obligatorily subcategorizes for a specific lexical item as an object in order to convey a certain (usually idiomatic) sense.* For example, in the following sentence the direct object, 'kafa' ('head' in English) in accusative case, is obligatory for the idiomatic usage:

(7) a. Adam kafayı yedi.
         `man   head+ACC eat+PAST+3SG`
         *'The man got mentally deranged.'*

The definiteness of the direct object adds a minor twist to the default order. If the direct object is an indefinite noun phrase, then it has to be immediately preverbal. This is due to the fact that, both the subject and the indefinite direct object have no surface case-marking that distinguishes them, so word order constraints come into play to force this distinction.

In order to present the flavor of word order variations in Turkish, we provide the following examples. These two sentences are used to describe the same event (i.e., have the same logical form), but they are used in different discourse situations. The first sentence presents constituents in a neutral default order, while in the second sentence, the time adjunct, 'bugün' (today), is the topic and the subject, 'Ahmet', is the focus:

(8) a. Ahmet bugün evden okula otobüsle 3 dakikada
         `Ahmet today home+ABL school+DAT bus+WITH 3 minute+LOC`
         *'Ahmet went from home to school by bus*

   gitti.
   `go+PAST+3SG`
   *in 3 minutes today.'*

  b. Bugün evden okula  otobüsle 3 dakikada Ahmet
    `today home+ABL school+DAT bus+WITH 3 minute+LOC Ahmet`
    *'It was Ahmet who went from home to school in 3 minutes*

    gitti.
    `go+PAST+3SG`
    *by bus today.'*

Although, sentences (b) and (c), in the following example, are both grammatical, (c) is not acceptable as a response to the question (a):

(9) a. Ali nereye  gitti?
    `Ali where+DAT go+PAST+3SG`
    *'Where did Ali go?'*

  b. Ali okula  gitti.
    `Ali school+DAT go+PAST+3SG`
    *'Ali went to school.'*

  c. \* Okula  Ali gitti.
    `school+DAT Ali go+PAST+3SG`
    *'It was Ali who went to school.'*

The word order variations exemplified by (9) are very common in Turkish, especially in discourse.

## 3.2.2 Existential Sentences

Existential sentences are sentences which have a verb denoting existence ( root word is 'var') or nonexistence (root word is 'yok'). Typical constituents of existential sentences are: poss-subj (possessor of subject), (expression of) time, (expression of) place, subject and verb. *Poss-subj* is separated from the subject noun phrase, because Turkish allows for the intervention of time and place adjuncts between the possessor and the remaining of the subject noun phrase. The possessor can also move to any position of the sentence independently of the subject. For example, both of the following Turkish sentences are grammatical, and they have the same logical form:

(10) a. Benim evde     iki  kitabım    var.
        I+GEN home+LOC two book+P1SG existent
        *'I have two books at home.'*

   b. Evde     iki  kitabım    var        benim.
      home+LOC two book+P1SG existent I+GEN
      *'At home, I have two books.'*

In the first sentence, the possessor of the subject noun phrase is at the sentence-initial (topic) position, whereas in the second sentence, the possessor of the subject noun phrase is at the post verbal (background) position.

In the absence of any control information, the constituents of existential sentences have the following default order:

   *poss-subj, time, place, subject, verb*

The verb is again obligatory like in predicative sentences. Some other constituents may not intervene between the subject and the verb. The location for the constituent which is the focus, is the position immediately preceding the subject. The locations for topic and background are again the sentence initial position and post verbal position, respectively.

### 3.2.3  Attributive Sentences

Attributive sentences are used in order to express some property of an entity (the subject of the sentence). This may be the location, quality, quantity, owner, order, etc. of the subject. The constituents of such sentences are: subject, pred-property, (expression of) time, and (expression of) place. The *pred-property* is the constituent conveying a property of the subject noun phrase.

In the absence of any control information, the constituents of existential sentences have the following default order:

*subject, time, place, pred-property*

The *pred-property* can be a specifier, a modifier, or a noun phrase (which can be in any case, except the accusative case). For example, in the following sentences, the *pred-property* is a specifying-relation, a possessor, a qualitative modifier, a quantitative modifier, an ordinal, a noun phrase in the nominative case, and a noun phrase in the ablative case, respectively:

(11) a. Kalemim    masada.
        `pencil+P1SG table+LOC`
        '*My pencil is on the table.*'

(12) a. Bu   kalem   benim.
        `this pencil I+GEN`
        '*This pencil is mine.*'

(13) a. Bu   kalem   kırmızı.
        `this pencil red`
        '*This pencil is red.*'

(14) a. Kalemlerin      sayısı         iki.
        `pencil+PLU+GEN number+P3SG two`
        '*The number of pencils is two.*'

(15) a. Bu   çocuk sınıfta       ikincidir.
        `this child classroom+LOC second+COPULA`
        '*This child has a rank of two in the classroom.*'

(16) a. Bu  bir köpektir.
     ```
     this a  dog+COPULA
     ```
     *'This is a dog.'*

(17) a. Gelişimiz        Ankara'dan.
     ```
     come+PART+P1PL Ankara+ABL
     ```
     *'Our coming is from Ankara.'*

## 3.3  Complex Sentences

Complex sentences are combinations of simple sentences (or complex sentences themselves) which are linked by either conjoining or various relationships like conditional dependence, cause-result, etc. An example complex sentence formed by the conjunction of two simple sentences is:

(18)   Kapıyı    açtım          ve  odaya    girdim.
       ```
       door+ACC open+PAST+1SG and room+DAT enter+PAST+1SG
       ```
       *'I opened the door and entered the room'*

The following sentences are also complex sentences formed by two simple sentences which are combined by conditional dependence (the first one) and cause-result relationship (the next two), respectively:

(19)   Kitabı    okursan              sorunun         cevabını
       ```
       book+ACC read+AOR+COND+3SG question+GEN answer+P3SG+ACC
       ```
       *'If you read the book, you will find the answer to your question.*

       bulacaksın.
       ```
       find+FUT+2SG
       ```

(20)   Sen    geldiğin           için        o     gitti.
       ```
       you   come+PART+P2SG  because   he   go+PAST+3SG
       ```
       *'He went because you came.'*

(21)  Gelmesini            istemediğimden             onu
      `come+INF+P3SG+ACC want+NEG+PART+P1SG+ABL he+ACC`
      *'Since I did not want him to come, I did not call him.'*

   çağırmadım.
   `call+NEG+PAST+1SG`

## 3.4  Noun Phrases in Turkish

Noun phrases are one of the fundamental components of natural language sentences. They are used to denote entities and events in the real world. They function in many roles, such as the subject or the object in a sentence or a sentential clause.

A noun phrase may consist of only one word, which can either be noun (or a simple modifier), or a pronoun, or it may consist of more than one word. The distinguished component of a noun phrase is called the *head* of the noun phrase. It can be specified, modified and/or classified by other constituents, referred to as *specifiers*, *modifiers* and *classifiers*, respectively.

The constituents of a Turkish noun phrase have an almost fixed order:

1. Set specifier,
2. Possessor,
3. Specifying relation,
4. Demonstrative specifier,
5. Quantifier,
6. Modifying relation,
7. Ordinal,
8. Quantitative modifier,

9. Qualitative modifier (zero or more),

10. Classifier, and

11. Head.

The first five of the above constituents are specifiers, and the next four are modifiers. All of these constituents are optional. In the following sections, we describe these constituents in detail.

As can be seen from this order, speficiers almost always precede modifiers and modifiers almost always precede classifiers,[4] which precede the head noun, although there are numerous exceptions. Also, within each group, word order variation is possible due to a number of reasons:

- The order of quantitative and qualitative modifiers may change: the aspect that is emphasized is closer to the head noun. For example, to denote "two red pencils", both of the following two noun phrases can be used in Turkish:

  (22) a. iki    kırmızı kalem  
         `two red     pencil`  
         *'two red pencils'*

      b. kırmızı iki   kalem  
         `red     two pencil`  
         *'two red pencils'*

  The indefinite singular determiner may also follow any qualitative modifier and immediately precede any classifier and/or head noun.

- Depending on the quantifier used, the position of the demonstrative specifier may be different. For example, in the first noun phrase below, the demonstrative specifier precedes the quantifier, whereas in the second one the quantifier precedes the demonstrative specifier:

---

[4]A classifier in Turkish is a nominal modifier which forms a noun–noun noun phrase, essentially the equivalent of *book* in forms like *book cover* in English. We use the term modifier for adjectival modifiers and not for nominal ones.

(23) a. bu    birkaç    kalem
       `this several pencil`
       *'these several pencils'*

   b. bütün o    kağıtlar
       `all   that paper+PLU`
       *'all those papers'*

This is a strictly lexical issue and not explicitly controlled by the feature structure, but by the information (stored in the lexicon) about the determiner used.

- The order of lexical and phrasal modifiers (e.g., corresponding to a postpositional phrase on the surface) may change, if positioning the lexical modifier before the phrasal one causes unnecessary ambiguity (i.e., the lexical modifier in that case can also be interpreted as a modifier of some internal constituent of the phrasal modifier). For example, the first noun phrase below has two interpretations. However, there is no such problem in the second one:

(24) a. iki kalemli    adam
       `two pencil+WITH man`
       *'two men with a pencil'*
       *'a man with two pencils'*

   b. kalemli    iki adam
       `pencil+WITH two man`
       *'two men with a pencil'*

So, phrasal modifiers always precede lexical modifiers and phrasal specifiers precede lexical specifiers, unless otherwise is specified, in which case punctuation needs to be used.

- A modifier may come after the classifier. For example in the first noun phrase below an ordinal modifier intervened between the classifier and the head, and in the second one, the intervening constituent is a qualitative modifier:

(25) a. futbol birinci ligi
    ```
    soccer first league+P3SG
    ```
    'first soccer league'

  b. Türkiye milli parkları
    ```
    Turkey national park+PLU+P3SG
    ```
    'national parks of Turkey'

- The possessor may scramble to a position past the head or even outside the phrase (to a background position), or allow some adverbial adjunct intervene between it and the rest of the noun phrase, *causing a discontinuous constituent*. For example, the possessor of the subject in the following sentence has moved to a background position:

(26) a. Kedisini gördün mü Ayşe'nin?
    ```
    cat+P3SG+ACC see+P2SG+QUES Ayşe+GEN
    ```
    'Did you see Ayşe's cat?'

Although we have included control information for scrambling the possessor to post head position, we have opted not to deal with either discontinuous constituents or long(er) distance scramblings as these are mainly used in spoken discourse.

- Furthermore, since the possessor information is explicitly marked on the head noun, sometimes the discourse does not require an overt possessor. For example, if the owner of the pencil is not to be emphasized, both of the following noun phrases can be used to denote "your pencil":

(27) a. senin kalemin
    ```
    you+GEN pencil+P2SG
    ```
    'your pencil'

  b. kalemin
    ```
    pencil+P2SG
    ```
    'your pencil'

In the following subsections, we present the constituents of a noun phrase in depth.

### 3.4.1 Specifiers

Specifiers are constituents of noun phrases which are used to distinguish the head noun out of a set of possible similar nouns in the context. The specifiers of a noun phrase are: the quantifier, the demonstrative specifier, the specifying relation, the possessor, and the set specifiers.

**The Quantifier**

Quantifiers are used to pick out the quantity of items denoted by the head noun. The Turkish **quantifiers** are: 'her' (every), 'biraz' (a little), 'bazı' (some), 'birkaç' ( a few), 'birçok' (many), 'bütün' (all), 'tüm' (all), 'kimi' (some) and the indefinite article 'bir' (a/an).

In Turkish, some quantifiers can only specify heads that are morphologically marked plural, and some can only specify heads that are singular. For example:

(28) a.  her    insan
         `every human`
         *'every human'*

   b. *  her    insanlar
         `every human+PLU`

(29) a.  bazı  masalar
         `some table+PLU`
         *'some tables'*

   b. *  bazı  masa
         `some table`

The countability of the head also plays an important role in the selection of the quantifier. Some quantifiers can only specify countable heads, whereas some others can only specify uncountable heads.

In Turkish, sometimes a demonstrative specifier and a quantifier can specify the same head. This is a property of the quantifier. For example:

(30) a. şu   birkaç   öğrenci
      `that several student`
     *'those several students'*

  b. * bu   bazı   kitaplar
      `this some book+PLU`

The order of the demonstrative specifier and the quantifier may also change, depending on the quantifier used. For example:

(31) a. bütün bu   kalemler
      `all   this pencil+PLU`
     *'all these pencils'*

  b. * birkaç şu   öğrenci
      `some  that student`

All of the above properties of quantifiers are coded in the lexicon.

The indefinite article 'bir' (a/an) can also be considered as a quantifier. With 'bir', the word order can change and it may occur between the qualitative modifiers and the classifier. The surface form of the indefinite article, 'bir', is also the surface form of the cardinal 'one'. If the word 'bir' is preceding a qualitative modifier in the surface form, it can either be an indefinite article or a cardinal, but if it is succeeding a qualitative modifier, then it is an indefinite article.

The presence of the indefinite article depends on the definiteness, specificity, and referentiality of the head noun. These are explained in detail in the following sections.

**The Demonstrative Specifier**

**Demonstrative specifiers** are used to point out items [16, p. 145]. Turkish demonstrative specifiers are: 'bu' (this), 'şu' (that), and 'o' (that). The demonstrative specifiers distinguish between the degrees of proximity to the speaker. 'bu' is used to point out items that are near the speaker. 'şu' and 'o' are used to point out items which are not near the speaker, but the items pointed out by 'şu' are closer (and possibly visible) to the speaker than the ones pointed out by 'o'.

**The Specifying Relation**

**Specifying relations** are used to pick out items by giving their relationship with other items. In Turkish, a specifying relation is a postpositional phrase formed by a noun phrase and one of the postpositions 'ait' (belonging to), 'dair' (about), etc. or an adjectival phrase, formed by the +ki relativizer from a singular noun phrase (with temporal or spatial location semantics) in the nominative case, or a noun phrase in the locative case.

If the specifying relation is a postpositional phrase, then the postposition gives the relationship of its argument noun phrase, with the head. For example, in the following noun phrase, the specifying relation is a postpositional phrase and the postposition gives a relation of ownership:

(32) a.  Ali'ye     ait          kitap
         `Ali+DAT belonging-to book`
         *'The book that belongs to Ali'*

If the postpositional phrase mentions a spatial location, it gets the `+ki` relativization suffix. For example:

(33) a.  hastaneden    önceki     ev
         `hospital+ABL before+REL house`
         *'house before the hospital'*

The specifying relation which is an adjectival phrase, mentions a spatial or temporal location. The following are the examples of noun phrases having an adjectival phrase as a specifying relation:

(34) a.  Ali'nin    evdeki           kitabı
         Ali+GEN home+LOC+REL book+P3SG
         *'Ali's book at home'*

(35) a.  dünkü             sınav
         yesterday+REL examination
         *'the exam yesterday'*

**The Possessor**

**Possessor** distinguishes an item by expressing its owner.[5] In Turkish, the possessor is a noun phrase with a genitive case marker. The agreement of the possessor should be the same as the possessive marker of the head noun, when it is present.[6] The information expressed by the possessor, can also be expressed by the possessive marker of the head alone, if such emphasis is not necessary in the context. The following two noun phrases, then, have almost the same semantics, but in the first one it is emphasized that 'the book is mine, as opposed to somebody else's', while the latter is neutral:

(37) a.  benim kitabım
         my    book+P1SG
         *'my book'*

   b.  kitabım
       book+P1SG
       *'my book'*

---

[5]This ownership includes any kind of possession. For example, in the noun phrase "John's book", John may also be the writer of the book.

[6]In general, the head noun need not have a possessive in Turkish, if there is a possessor. For example, the following is also a grammatical noun phrase:
(36) a.  benim kitap
         my    book
         *'my book'*
But, such forms are used very infrequently, and should probably not be dealt with at this point.

Therefore, (37b) cannot be used in the answer to (38a) in the following discourse, where the owner of the book should be emphasized:

(38) a.  Kimin kitabı    kalın?
         `whose book+P3SG thick`
         *'Whose book is thick?'*

   b.  Benim kitabım    kalın.
       `I+GEN book+P1SG thick`
       *'My book is thick.'*

   c. * Kitabım    kalın.
        `book+P1SG thick`

The possessor is the only constituent in a noun phrase, which can move from its position to a position past the head. For example, both of the following sentences mention that 'my house is beautiful', but they convey different information, regarding focus, topic and background:

(39) a. Benim evim        güzeldir.
        `I+GEN house+P1SG beautiful+COPULA`
        *'My house is beautiful.'*

   b. Evim        güzeldir        benim.
      `house+P1SG beautiful+COPULA I+GEN`
      *'My house is beautiful.'*

The possessor of the subject is backgrounded in the latter.

### The Set Specifier

If a noun phrase is specified by a **set specifier**, then the noun phrase denotes the members of some set, which have some distinguishing identity or property, or which are of some quantity. The set specifier is a noun phrase which is semantically plural and is in the ablative case. For example:

(40) a. akrabalardan amcamın kızı
     `relative+PLU+ABL uncle+P1SG+GEN daughter+P3SG`
     'my uncle's daughter among relatives'

If the head, which is specified by a set specifier, is the same as that of the set specifier, then it drops. In such cases, one of the modifiers of the head act as the head. For example:

(41) a. adamlardan bu ikisi
     `man+PLU+ABL this two+P3SG`
     'these two among the men'

   b. kazaklardan evdeki ikisi
     `pullover+PLU+ABL home+LOC+REL two+P3SG`
     'among the pullovers, the two at home'

   c. kitaplardan en kalını
     `book+PLU+ABL most thick+P3SG`
     'the thickest one among the books'

### 3.4.2 Modifiers

Modifiers are constituents of noun phrases which give information about the properties of the concept denoted by the head noun, or about the relations of its properties with properties of other concepts. The modifiers of a noun phrase are: the modifying relation, the ordinal, the quantitative modifier, and the qualitative modifier.

#### The Modifying Relation

**Modifying relation** gives information about the properties of a concept. This property can also be given in comparison with another concept. In Turkish, a modifying relation can either be a postpositional phrase, like the specifying relation, or a noun phrase followed by one of the suffixes `+DAn` (ablative, made of), `+lH` (with), `+sHz` (without), `+DA` (locative, made on), or `+lHk` (of).

The postpositional phrase gives the property in comparison with another item. It is formed by a noun phrase and one of the postpositions 'gibi' (like), 'kadar' (as much as), 'önce' (before), etc. For example:

(42) a. at    gibi köpek
     `horse like dog`
     *'dog like a horse'*

The case of the noun phrase should match the subcategorization requirement of the postposition used. If the head of the noun phrase the whole noun phrase is compared to is the same as the head of the whole noun phrase, then it may be eliminated from the surface form. For example:

(43) a. ? evdeki        elmalar    kadar       elma
         `home+LOC+REL apple+PLU as many as apple`
         *'as many apples as there are at home'*

   b.   evdeki       kadar       elma
        `home+LOC+REL as many as apple`
        *'as many apples as there are at home'*

The modifying relation can also be a noun phrase followed by one of the suffixes:

1. `+DAn` (ablative case marking indicating "made of" relationship),

   (44)   tahtadan masa
          `wood+ABL table`
          *'table made of wood'*

2. `+lH` (adjective derivation suffix indicating "with"),

   (45)   örtülü    masa
          `cover+WITH table`
          *'table with cover'*

3. `+sHz` (adjective derivation suffix indicating "without"),

(46)    örtüsüz      masa
        `cover+WITHOUT table`
        *'table without cover'*

4. `+DA` (locative marker indicating relationship "made-on"),

(47)    kiremitte   şiş
        `brick+LOC kebap`
        *'kebap (cooked) on brick'*

5. or `+lHk` (adjective derivation suffix indicating "has measurable property" relationship),

(48)    iki  kiloluk   karpuz
        `two kilo+OF watermelon`
        *'a watermelon weighing two kilos'*

If the suffix is `+DAn`, then the head of the noun phrase must have the semantic property of denoting a material, though this may be relaxed.

### The Ordinal

An **ordinal** expresses the order of an item. In Turkish, an ordinal modifier can be one of: ilk (first), birinci (first), ikinci (second), üçüncü (third), etc., and sonuncu (last). Although the words 'ilk' and 'birinci' seem to have the same semantics, they have a slightly different behaviour.

In Turkish, there can be an intensifier which modifies the ordinal. This is the adverb 'en' (most). For example:

(49)    en     sonuncu masa
        `most last    table`
        *'the (most) last table'*

But this adverb can only intensify the ordinals 'birinci' and 'sonuncu'.[7]

---

[7] Note that 'birinci' can take an adverbial intensifier, whereas 'ilk' cannot.

**The Quantitative Modifier**

A **quantitative modifier** expresses the quantity of the referent denoted by the head noun. It may be, one of:

1. a cardinal:

    (50)   evdeki       iki  masa
           `home+LOC+REL two table`
           *'two tables at home'*

2. a range:

    (51)   bu   üç   beş  kalem
           `this three five pencil`
           *'these 3 to 5 pencils'*

3. an adjective that expresses a fuzzy quantity (for example: az or çok):

    (52)   çok elma
           `lot apple`
           *'lots of apples'*

    Note that this adjective can also be followed by one of the words 'miktarda' or 'sayıda', if the head is uncountable and countable, respectively. This usage is more formal.

4. a noun phrase, where the head is a container or a measure noun modified by a quantitative modifier of type cardinal or range. If the head is a container, then the noun phrase specifying the quantity may also be followed by the word 'dolusu', meaning 'full of'. For example:[8]

    (53) a. üç     bardak su
            `three glass  water`
            *'three glasses of water'*

       b. üç     bardak dolusu   su
            `three glass full of water`
            *'three glassfuls of water'*

---

[8]Note that (48) and (53c) are different.

       c. iki   kilo   karpuz
         `two kilo watermelon`
         *'two kilos of watermelon'*

In Turkish, a head cannot be modified by a quantitative modifier and specified by a quantifier at the same time, though some quantifiers allow for the presence of a cardinal. For example:

(54) a.   her    iki   kalem
       `every two pencil`
       *'both pencils'*

    b. * bazı  iki   masalar
       `some two table+PLU`

The cardinal may specify the quantity as an upper or a lower limit. For example:

(55) a. en    az     iki   elma
      `most little two apple`
     *'at least two apples'*

    b. en    çok   iki   kalem
      `most much two pencil`
     *'at most two pencils'*

The limit may also be given as a range as well as a cardinal. For example:

(56) a. en    az    iki   üç     kişi
      `most less two three person`
     *'at least two to three people'*

The cardinal which specifies a quantity may be followed by one of the words 'adet', 'tane' and 'parça' (all meaning 'piece'), but if the quantitative modifier is also a noun phrase and its head is a measure noun, then these words cannot be present between the cardinal and the measure noun in the surface form. For example:

(57) a.  iki  elma
         `two apple`
         'two apples'

   b.  iki  adet  elma
       `two piece apples`
       'two apples'

   c. * iki  adet  kilo  elma
        `two piece kilo apple`

If the head is modified by a quantitative modifier which is a cardinal greater than one, or a quantitative modifier which is a range, then this noun phrase is semantically plural. But, in Turkish, such a noun phrase is not morphologically marked plural at the same time.

#### The Qualitative Modifiers

A **qualitative modifier** expresses qualitative properties of a concept. There may be any number of qualitative modifiers modifying the head. In Turkish, a qualitative modifier is an adjective phrase, which is formed from ordinary adjectives. For example:

(58)  büyük sarı   kapı
      `big   yellow door`
      'big yellow door'

An adjective in the adjective phrase can also be modified by an adverb or a postpositional phrase which function as intensifiers. In the noun phrase in (59a), the qualitative adjective 'büyük' (big) is modified by an adverb, whereas in the noun phrase in (59b), it is modified by a postpositional phrase [23, p. 160]:

(59) a. en  büyük masa
    most big  table
    'the biggest table'

  b. at   kadar büyük bir köpek
    horse as  big   a  dog
    'a dog as big as a horse'

The order of quantitative and qualitative modifiers modifying the head is not fixed. The one that is emphasized is closer to the head in the surface form. For example, in the first phrase below, the fact that 'the tables are big' is emphasized, while in the second one, the fact that 'there are two tables' is emphasized:

(60) a. iki  büyük masa
    two big   table
    'two big tables'

  b. büyük iki  masa
    big   two table
    'two big tables'

### 3.4.3 Classifiers and the Head

The **head** of a noun phrase is either a proper noun, common noun or a pronoun. Pronouns cannot take any specifiers, modifiers, or classifiers. Nouns can further be classified by a set of **classifiers**. A classifier can also be a common noun, or it can itself be a noun phrase which can only have a classifier, classifying the head and/or modifiers modifying the head. Furthermore, the head of a noun phrase can also be a noun phrase having only a classifier classifying the head. For example, in the following noun phrase, the head and the classifier are both noun phrases, each having a classifier:

(61)   kredi  kartı    komisyon   oranı
    credit card+P3SG commission rate+P3SG
    'credit card commission rate'

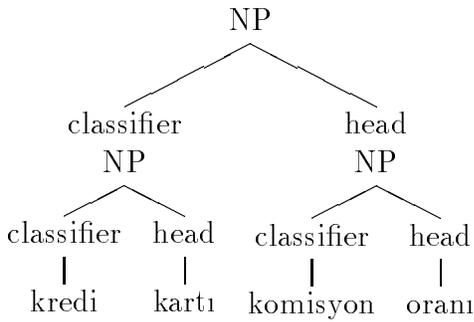

Figure 3.1: The structure of the noun phrase in (61)

The tree in Figure 3.1 is given to demonstrate the structure of this noun phrase.

A modifier may intervene between the classifier and the head. The following noun phrase is an example to such a form:[9]

(62) a. Dışişleri       eski Bakanı
     `foreign affairs old minister+P3SG`
     *'old Minister of Foreign Affairs'*

The head of the noun phrase can sometimes drop. This is usually the case, when the head is already introduced into the discourse, and its other distinguishing properties also need to be introduced. In such a case, the classifier of the noun phrase, if present, drops, too. A modifier or a specifier (only the possessor, the quantifier, or the specifying relation) of the noun phrase, which is immediately preceding the head in the normal surface form substitutes for the head and gets the case, number and possessive markings of the head. If this element is possessor, then it gets the suffix +ki, and then only the number and case markings. For example:

(63) a. Kimin kitabı    kalın?
     `whose book+P3SG thick`
     *'Whose book is thick?'*

---
[9]See also earlier examples.

    b. Ayşe'nin kitabı  kalın.
       `Ayşe+GEN book+P3SG thick`
       *'Ayşe's book is thick.'*

    c. Ayşe'ninki  kalın.
       `Ayşe+GEN+REL thick`
       *'That of Ayşe is thick'*

If the element which is substituting for the head, is a modifying element other than the possessor, it gets both of the number, case and possessive markings of the head. For example:

(64) a. Hangi kalemi  istiyorsun?
       `which pencil+ACC want+PRG+2SG`
       *'Which pencil do you want?'*

    b. Kırmızıyı istiyorum.
       `red+ACC  want+PRG+1SG`
       *'I want the red (one).'*

### 3.4.4  Definiteness, Specificity, and Referentiality

Three main distinctions that underlie the interpretation of a noun phrase are *definiteness, specificity,* and *referentiality.* These are some of the factors which contribute in the determination of the case of the noun phrase and in the determination of the presence of the indefinite article.

- If it is possible for the hearer to build an unambiguous link between a noun phrase, and an entity or a group of entities, then the noun phrase is **definite**. The underlined noun phrase in the following sentence is an example to a definite noun phrase:

  (65)    Mary dropped <u>the book</u> on the table.

  The underlined noun phrases in the following sentences are examples of definite and indefinite noun phrases in Turkish, respectively:

(66) a. Ahmet <u>tarih kitabını</u> okuyor.
    Ahmet history book+P3SG+ACC read+PRG+3SG
    'Ahmet is reading his history book.'

b. Ali'ye <u>bir kitap</u> verdim.
    Ali+DAT a book give+PAST+1SG
    'I gave a book to Ali.'

- If the entity that the noun phrase is linked to is contextually bound (i.e. element of the universe of discourse), then the noun phrase is **specific** [13]. The underlined noun phrase in the following sentence is an example to a noun phrase, which is definite and specific:

(67)   Three cats entered the kitchen. <u>They</u> ate the cake.

The underlined noun phrases in the following sentences are examples of specific and non-specific noun phrases in Turkish, respectively:

(68) a. Ali <u>kitap</u> okumayı sevmiyor.
    Ali book read+INF+ACC like+NEG+PRG+3SG
    'Ali doesn't like reading book.'

b. <u>Kitabını</u> okulda bıraktı.
    book+P3SG+ACC school+LOC leave+PAST+3SG
    'He left his book at school.'

- A noun phrase is **referential**, if there is a particular object or a set of objects within the relevant universe of discourse, that the noun phrase refers to [13]. The underlined noun phrases in the following sentence are examples to noun phrases in English and Turkish, which are used referentially:

(69)   Mary went to <u>the cinema</u> to watch 'Blues Brothers'.

(70)   <u>Kütüphaneye</u> kitap okumak için gittim.
    library+DAT book read+INF for go+PAST+1SG
    'I went to the library for reading a book.'

A noun phrase is **non-referential**, if there is no such object and the noun phrase denotes a concept without pointing out any particular individual [13]. The underlined noun phrases in the following sentences are examples to noun phrases in English and Turkish, which are used non-referentially:

(71)  Jim went to high-school in Paris.

(72)  Ali <u>okula</u>    altı yaşında    başladı.
      `Ali school+DAT six age+P3SG+LOC start+PAST+3SG`
      'Ali started school when he was six years old.'

Definiteness, specificity, and referentiality of noun phrases can be signaled by various strategies like morphological marking, word order, stress and context. The marking of noun phrases in Turkish and how this correlates with definiteness, specificity, and referentiality are summarized [4] in the following table:[10]

|  |  | Referential |  | Non-referential |
|---|---|---|---|---|
|  | definite | Indefinite |  |  |
|  |  | Specific | Non-specific |  |
| Subject sg. | __-$\phi$ | bir __ | bir __ | __-$\phi$ |
| pl. | __+lAr |  |  | __+lAr |
| Object sg. | __+yH | bir __+yH | bir __(+yH) | __+$\phi$ |
| pl. | __+lAr+yH |  |  | __+lAr+yH |
| Oblique sg. | __+CASE | bir __+CASE | bir __+CASE | __+CASE |
| pl. | __+lAr+CASE |  |  | __+lAr+CASE |

## 3.4.5 Multiple Specifiers and Modifiers

There may be more than one modifier or specifier of one kind, modifying or specifying the head, or a noun phrase may be a conjunction (or a disjunction) of more than one noun phrases, forming a list. Examples to these can be given as follows:

---

[10]+<u>yH</u> is the accusative case morpheme; the glide [y] is omitted when the accusative ending is attached to a word ending in a consonant, H denotes a high vowel {ı,i,u,ü} resolved according to vowel harmony rules.

+<u>lAr</u> is the plural morpheme (A denotes {a,e}). Parentheses indicate optionality.

(73) a. tahtadan, örtüsüz       masa
       `wood+ABL cover+WITHOUT table`
       *'a table made of wood, without a cover'*

   b. tahtadan büyük ve yeşil masa
       `wood+ABL big   and green table`
       *'big and green table made of wood'*

   c. ihtiyar adam ve deniz
       `old   man  and sea`
       *'the old man and the sea'*

   d. iki veya üç araba
       `two or  three car`
       *'two or three cars'*

If the elements of this list are modifying relations or qualitative modifiers modifying a head, then there may be a comma between the elements. Instead of a comma, there may also be a conjunction or a disjunction between the last two elements. If the elements of this list are noun phrases, then there should be a disjunction or a conjunction between the last two elements.

## 3.5 Sentential Clauses

Sentential clauses correspond to either:

- full sentences with non-finite or participle verb forms which act as noun phrases in either argument or adjunct roles, or

- gapped sentences with participle verb forms which function as modifiers of noun phrases (the filler of the gap).

The former non-gapped forms in Turkish can be further classified into those representing *acts*, *facts* and *adverbials*. Figure 3.2 exhibits the classification of sentential clauses.

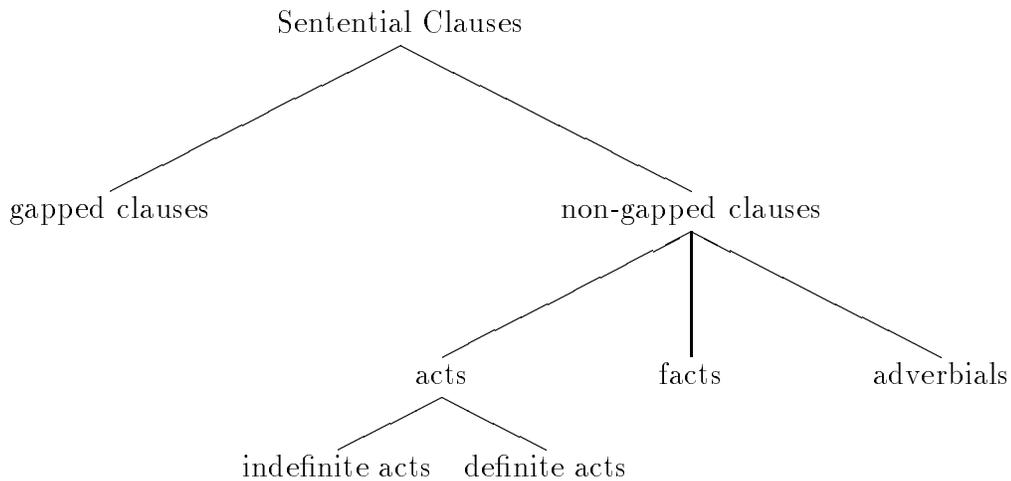

Figure 3.2: The classification of sentential clauses

Sentential arguments of verbs adhere to the same morphosyntactic constraints as the nominal arguments. For example, the participle of, say, a clause that acts as a direct object is case-marked accusative, just as the nominal one would be. The subject and the direct object of the following sentence are both sentential clauses, and the direct object is case marked accusative because of the reason above.

(74) Ali'nin buraya gelmesi bizim işi bitirmemizi
```
Ali+GEN here   come+INF+P3SG we+GEN job+ACC finish+INF+P1PL+ACC
```
'Ali's coming here made us finish the job easier.'

kolaylaştırdı.
```
make_easy+PAST+3SG
```

In the following subsections, acts, facts, adverbials, and gapped sentential clauses will be presented in detail.

## 3.5.1 Acts

The sentential clauses representing acts are classified into two: those representing *indefinite acts* and *definite acts*. The verbs of the sentential clauses representing an indefinite act end with the suffix `+mAk` (`+mak` or `+mek`). For example, the direct object of the following sentence, 'kitap okumak' ('to read a book' in English), is a sentential clause representing an indefinite act.

(75)    Kitap okumak   istiyorum.
        `book  read+INF want+PRG+1SG`
        *'I want to read a book.'*

The verbs of the sentential clauses representing a definite act take the suffix `+mA` (`+ma` or `+me`) or `+Hş` (`+iş`, `+ış`, `+uş`, or `+üş`) and a possessor suffix agreeing with the subject of the sentential clause. The direct object of the following sentence, 'gelmesi' ('his coming' in English), is an example to a sentential clause representing a definite act:

(76)    Gelmesini              istedim.
        `come+INF+P3SG+ACC want+PAST+1SG`
        *'I wanted him to come.'*

### 3.5.2  Facts

Sentential clauses which correspond to full sentences representing facts have participle verb forms. Their verbs can be in one of the two forms, depending on the time of the event denoted by the sentential clause. If that time precedes the time of the event expressed by the verb of the main sentence, or is concurrent, then the verb of the sentential clause takes a `+dHk` suffix, otherwise it takes a `+yAcAk` suffix. Furthermore, the verb of the sentential clause takes a possessor suffix, which agrees with the subject of the sentential clause. The direct objects of the following sentences are examples to such forms, respectively:

(77)    Kitap okuduğumu            gördü.
        `book  read+PART+P1SG+ACC see+PAST+3SG`
        *'He saw that I am reading a book.'*

(78)   Geleceğini              zannettim.
       come+PART+P3SG+ACC think+PAST+1SG
       'I thought that he will come.'

### 3.5.3  Adverbials

The verb forms of the sentential clauses which represent adverbials depend on the role and semantics of the sentential clause. The suffix that the verb gets can be +ArAk for a manner adverbial, +ken for a manner or a time adverbial, +yIncA for a time adverbial, etc. The manner adjunct of the following sentence is an example manner adverbial, and the time adjunct of the second sentence is an example temporal adverbial:

(79) a. [Koşarak] odaya    girdi.
        run+ADVB room+DAT enter+PAST+3SG
        'He entered the room running.'

    b. [Onu    buraya   gelirken]          gördüm.
        he+ACC here+DAT come+AOR+ADVB see+PAST+1SG
        'I saw him while he was coming here.'

A list of the suffixes for making adverbials is given in Appendix A.

### 3.5.4  Gapped Sentential Clauses

Gapped sentential clauses correspond to gapped sentences with participle verb forms (relative clauses) and they function as a modifier of the noun phrase, which is the filler of the gap. There are two strategies of relativization: *subject participle* (the gapped constituent is the subject of the clause) and *object participle* (the gapped constituent is anything other than the subject) [1, 6]. These differ in morphological markings on the verb and the subject of the clause and in the presence of an agreement between the subject and the verb of the clause. The selection of morphological markings depends on the thematic role of the gap, voice and transitivity of the verb of the sentential clause, and the specificity of the subject of the sentential clause. The suffix that the verb of the sentential clause gets can be: `+yAn, +yAcAk, +mHş` alone, or `+dHk, +yAcAk` plus a possessive suffix. The table given in Appendix B gives the suffix in different situations and an example noun phrase corresponding to each form.

In subject participles, since the subject is gapped, it is not marked genitive, and there is no agreement between the verb and the subject. The direct object of the following sentence is an example to subject participles:

(80)  [___$_i$ Odaya    giren]     adamı$_i$ biliyorum.
        room+DAT enter+PART man+ACC know+PRG+1SG
      '*I know the man$_i$ who$_i$ entered the room.*'

In object participles, the subject of the clause is marked genitive and the verb is marked with a possessive suffix (agreeing with the subject). The direct object of the following sentence is an example to object participles:

(81)  [Onun ___$_i$ verdiği]       kitabı$_i$  okudum.
       he+GEN    enter+PART+P3SG book+ACC read+PAST+1SG
      '*I read the book$_i$ (that$_i$) he gave me.*'

In Turkish, unbounded relativization, relativization in embedded clauses, is also possible. The direct object of the following sentence is an example to unbounded dependencies:

(82)  Adam [[[___$_i$ okumak] istediğini]         söylediği]    kitabı$_i$
      man          read+INF want+PART+P3SG+ACC say+PART+P3SG book+ACC
      '*The man didn't read the book that he told he wanted to read.*'

      okumadı.
      read+NEG+PAST+3SG

The details of relativization in Turkish are investigated by Barker *et.al.* [1] and Güngördü [6].

# Chapter 4

# Generation of Turkish Sentences

This chapter describes the generation of Turkish sentences. In the first section, we explain the architecture of our generator. Then, we present a comparison of our work with other related works.

## 4.1 The Architecture of the Generator

### 4.1.1 Approach

In order to generate Turkish sentences of varying complexity, we have designed a recursively structured finite state machine which can also handle the changes in constituent order. Our implementation environment is the GenKit system [22], developed at Carnegie Mellon University–Center for Machine Translation. Morphological realization has been implemented using an external morphological analysis/generation component, developed using XEROX Two Level Tools, which performs concrete morpheme selection and handles morphographemic processes.

The generation process gets as input a feature structure representing the content of the sentence where all the lexical choices have been made, then produces as output the surface form of the sentence. The feature structures for sentences are represented using a case-frame representation. The fact that sentential arguments of verbs adhere to the same morphosyntactic constraints as the nominal arguments enables a nice recursive embedding of case-frames of similar general structure to be used to represent sentential arguments.

### 4.1.2 Simple Sentences

In this section, we will explain the generation of predicative, existential, and attributive sentences. Our input feature structures and finite state machines for giving the outlines of predicative, existential, and attributive sentences differ slightly. Therefore, for each kind of simple sentence, we present the case-frame and the corresponding finite state machine.

### Predicative Sentences

We use the following general case-frame feature structure to encode the contents of a predicative sentence:

$$\begin{bmatrix}
\text{S-FORM} & \text{infinitive/adverbial/participle/finite} \\
\text{CLAUSE-TYPE} & \text{predicative} \\
\text{VOICE} & \text{active/reflexive/reciprocal/passive/causative} \\
\text{SPEECH-ACT} & \text{imperative/optative/necessitative/wish/} \\
& \text{interrogative/declarative} \\
\text{QUES} & \begin{bmatrix} \text{TYPE} & \text{yes-no/wh} \\ \text{CONST} & \text{list-of(subject/dir-obj/etc.)} \end{bmatrix} \\
\text{VERB} & \begin{bmatrix} \text{ROOT} & \text{verb} \\ \text{POLARITY} & \text{negative/positive} \\ \text{TENSE} & \text{present/past/future} \\ \text{ASPECT} & \text{progressive/habitual/etc.} \\ \text{MODALITY} & \text{potentiality} \end{bmatrix} \\
\text{ARGUMENTS} & \begin{bmatrix} \text{SUBJECT} & \textit{c-name} \\ \text{DIR-OBJ} & \textit{c-name} \\ \text{SOURCE} & \textit{c-name} \\ \text{GOAL} & \textit{c-name} \\ \text{LOCATION} & \textit{c-name} \\ \text{BENEFICIARY} & \textit{c-name} \\ \text{INSTRUMENT} & \textit{c-name} \\ \text{VALUE} & \textit{c-name} \end{bmatrix} \\
\text{ADJUNCTS} & \begin{bmatrix} \text{TIME} & \textit{c-name} \\ \text{PLACE} & \textit{c-name} \\ \text{MANNER} & \textit{c-name} \\ \text{PATH} & \textit{c-name} \\ \text{DURATION} & \textit{c-name} \end{bmatrix} \\
\text{CONTROL} & \begin{bmatrix} \text{IS} & \begin{bmatrix} \text{TOPIC} & \text{constituent} \\ \text{FOCUS} & \text{constituent} \\ \text{BACKGR} & \text{constituent} \end{bmatrix} \\ \text{ES} & \begin{bmatrix} \text{EVEN} & \text{constituent/poss-constituent} \\ \text{TOO} & \text{constituent/poss-constituent} \\ \text{QUES} & \text{constituent/poss-constituent} \end{bmatrix} \end{bmatrix}
\end{bmatrix}$$

We use the information given in the CONTROL | IS feature to guide our grammar in generating the appropriate sentential constituent order. This information is exploited by a right linear grammar (recursively structured nevertheless) to generate the proper order of constituents at every sentential level (including embedded sentential clauses with their own information structure). The simplified

outline of this right linear grammar is given as a finite state machine. The recursive behavior of this finite state machine comes from the fact that the individual argument or adjunct constituents can also embed sentential clauses. The details of the feature structures for sentential clauses are very similar to the structure for the case-frame. Thus, when an argument or adjunct, which is a sentential clause, is to be realized, the clause is recursively generated by using the same set of transitions.

The outline of a grammar for generating predicative sentences can be given by a recursive finite state machine (FSM). Before proceeding to this FSM, in order to ease the understanding of it, a simpler example in a simpler domain will be given. In this simple domain, the only arguments of the verb are subject and direct object, there is no adjunct, and the only constituent of the information structure is the focus. The default word order is:

*subject, direct object, verb.*

The constituent which is to be emphasized, the focus, moves to the immediately preverbal position. So, in the following example, the first sentence is in the default order, and in the second sentence, the subject, 'Ali', is the focus:

(83) a. Ali  topu     attı.             (Default Ord.)
    `Ali ball+ACC throw+PAST+3SG`
    *'Ali threw the ball.'*

   b. Topu    Ali  attı.             ('Ali' is focus)
    `ball+ACC Ali throw+PAST+3SG`
    *'It was Ali, who threw the ball.'*

The outline of a grammar for this simple domain can be given with the finite state machine in Figure 4.1.

In this finite state machine, the transition from the initial state to state 1, labeled **Subject** generates the subject, if it is defined and it is not the focus. Otherwise, a **NIL** transition is done, which generates an empty string. Then, the transition from state 1 to state 2, labeled **Dir-obj**, generates the direct object, if it is defined. The transition from state 2 to state 3, labeled **Subject**, generates the subject if it is the focus. Finally, the verb is generated with the transition from state 3 to the final state, labeled **verb**.

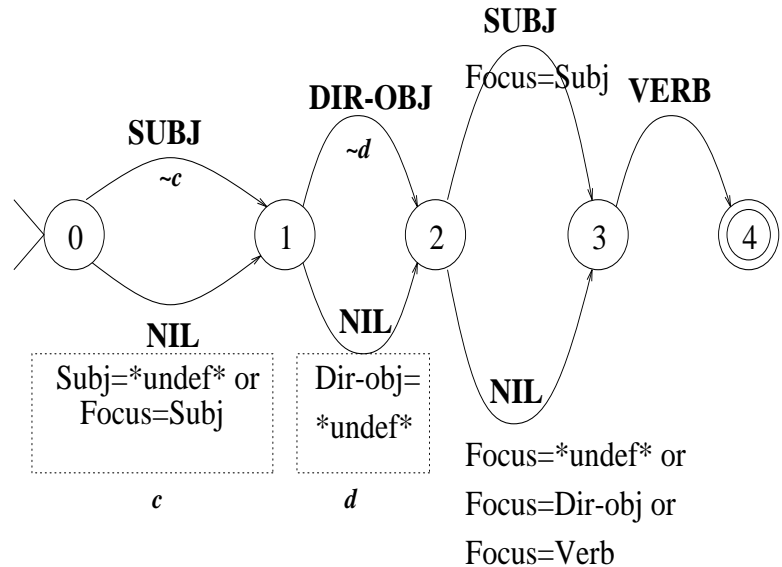

Figure 4.1: The finite state machine for giving the outline of a grammar for the simple domain.

Extending this simple domain to cover all the arguments of the verb, adjuncts, and information structure constituents, we get the finite state machine given in Figures 4.2 and 4.3 to give an outline of our grammar for generating predicative sentences.

In this finite state machine, transitions are labeled by constraints and constituents (shown in bold face along a transition arc) which are generated when those constraints are satisfied. If any transition has a **NIL** label, then no surface form is generated for that transition. The transitions from state 0 to state 1 generate the constituent which is the topic. If the topic is not defined, then the **NIL** transition is taken, which generates an empty string. The other transitions, from state 1 to state 14 generate the constituents in the default order. The transitions from state 14 to 15 generate the constituent which is the focus, and the constituent which is the background is generated by the transitions from state 17 to the final state.

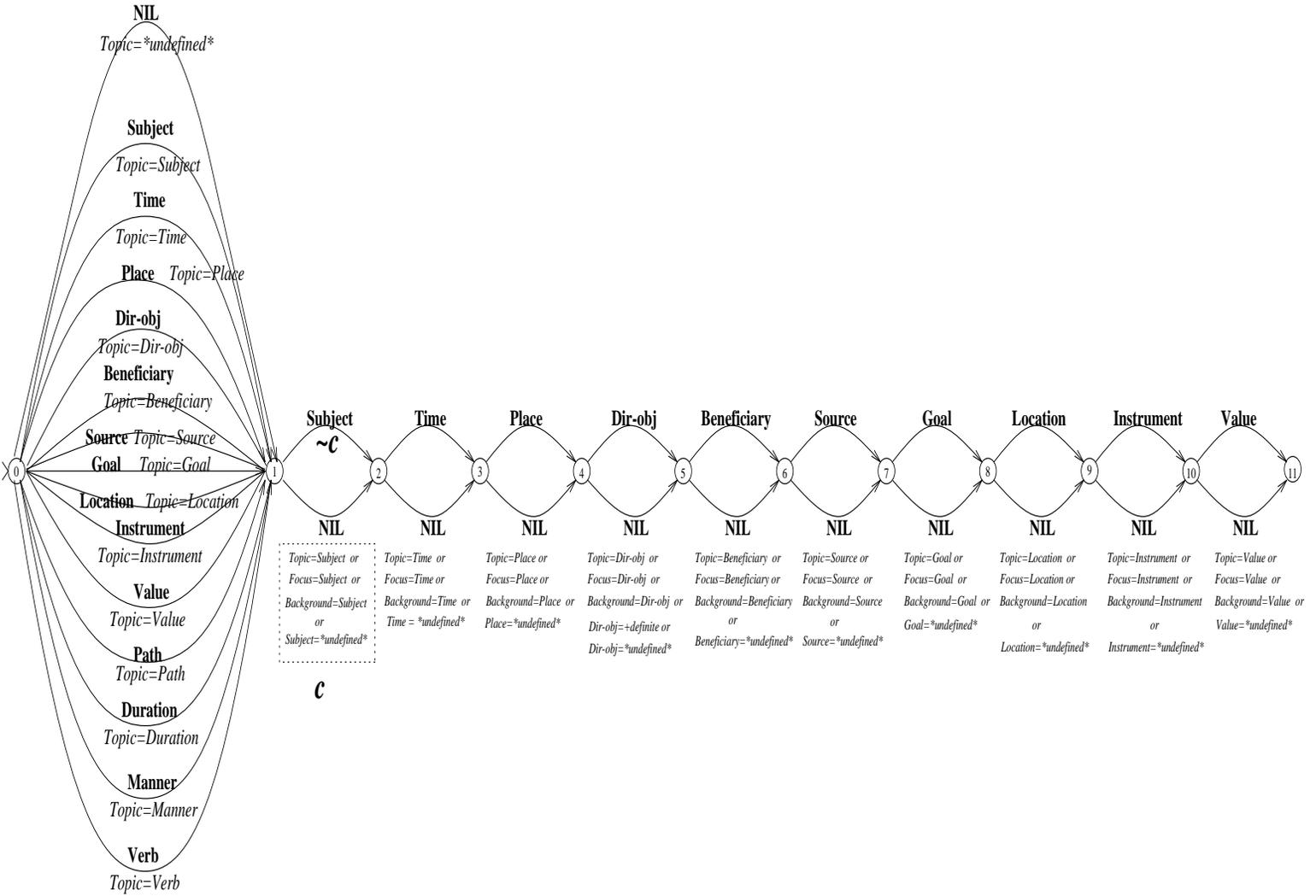

Figure 4.2: The finite state machine for predicative sentences (Part I).

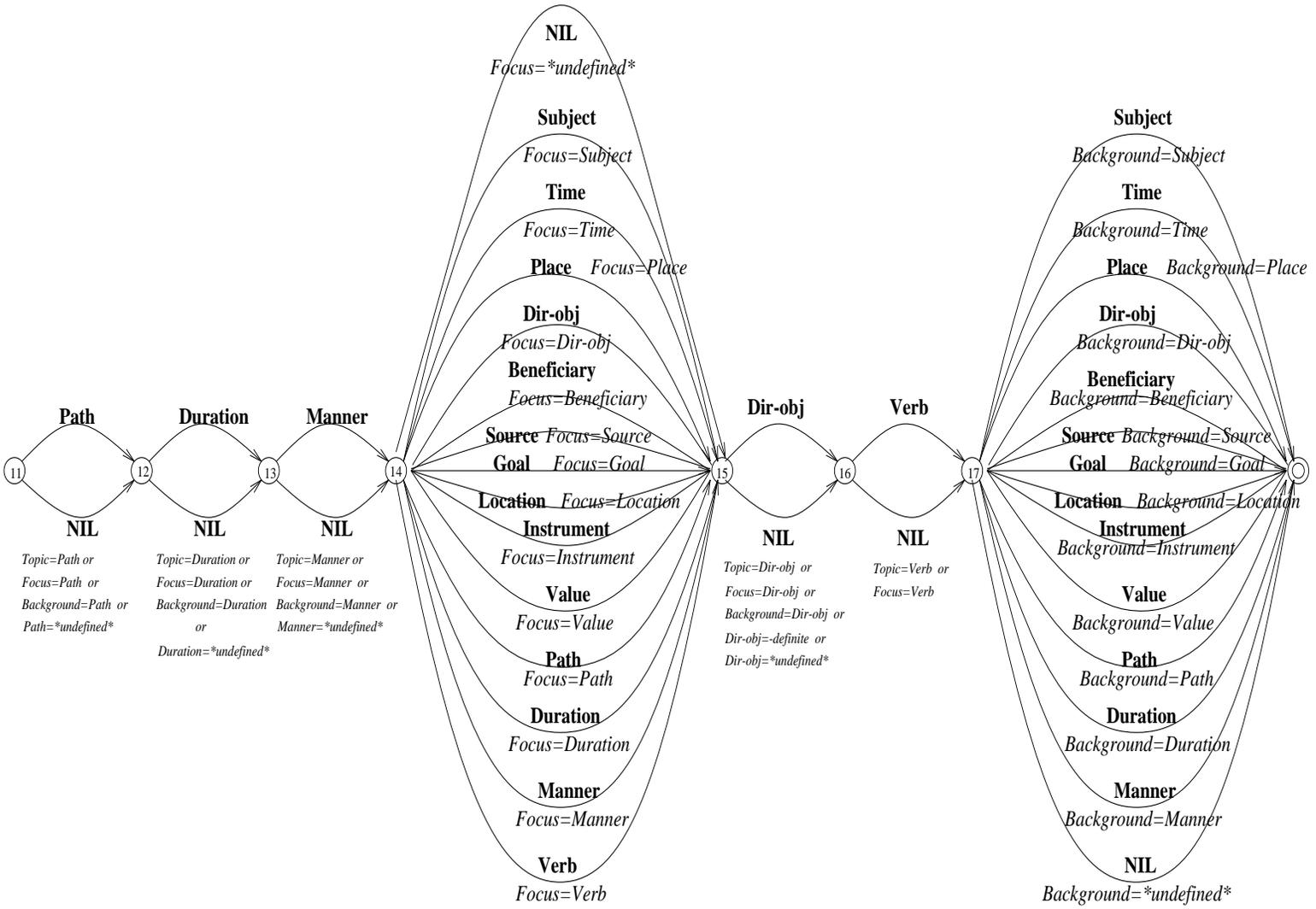

Figure 4.3: The finite state machine for predicative sentences (Part II).

When the following case-frame is sent to the generator, since it has no CON-TROL feature, and so no information structure, the constituents are generated in the default order, which is the sentence below the case-frame.

$$\begin{bmatrix} \text{S-FORM} & \text{finite} \\ \text{CLAUSE-TYPE} & \text{predicative} \\ \text{VOICE} & \text{active} \\ \text{SPEECH-ACT} & \text{declarative} \\ \text{VERB} & \begin{bmatrix} \text{ROOT} & \#\text{bırak} \\ \text{SENSE} & \text{positive} \\ \text{TENSE} & \text{past} \\ \text{ASPECT} & \text{perfect} \end{bmatrix} \\ \text{ARGUMENTS} & \begin{bmatrix} \text{SUBJECT} & \{\text{Ahmet}\} \\ \text{DIR-OBJ} & \{\text{kitap}\} \\ \text{LOCATION} & \{\text{masa}\} \end{bmatrix} \\ \text{ADJUNCTS} & \begin{bmatrix} \text{TIME}\{\text{dün}\} \end{bmatrix} \end{bmatrix}$$

(84)  Ahmet dün       kitabı    masada    bıraktı.
      `Ahmet  yesterday book+ACC table+LOC leave+PAST+3SG`
      *'Ahmet left the book on the table yesterday.'*

Considering the content, the following case-frame is the same as the previous one. However, this case-frame has a CONTROL feature, so an information structure, which expresses that expression of time is topic, subject is focus, and location is background. Therefore, the sentence generated when this case-frame is sent to the generator is not in the default order, which is the sentence below this case-frame. Figure 4.4 demonstrates the transitions done on the finite state machine to generate this sentence.

$$\begin{bmatrix} \text{S-FORM} & \text{finite} \\ \text{CLAUSE-TYPE} & \text{predicative} \\ \text{VOICE} & \text{active} \\ \text{SPEECH-ACT} & \text{declarative} \\ \text{VERB} & \begin{bmatrix} \text{ROOT} & \text{\#bırak} \\ \text{SENSE} & \text{positive} \\ \text{TENSE} & \text{past} \\ \text{ASPECT} & \text{perfect} \end{bmatrix} \\ \text{ARGUMENTS} & \begin{bmatrix} \text{SUBJECT} & \{\text{Ahmet}\} \\ \text{DIR-OBJ} & \{\text{kitap}\} \\ \text{LOCATION} & \{\text{masa}\} \end{bmatrix} \\ \text{ADJUNCTS} & \begin{bmatrix} \text{TIME}\{\text{dün}\} \end{bmatrix} \\ \text{CONTROL} & \begin{bmatrix} \text{IS} & \begin{bmatrix} \text{TOPIC} & \text{time} \\ \text{FOCUS} & \text{subject} \\ \text{BACKGROUND} & \text{location} \end{bmatrix} \end{bmatrix} \end{bmatrix}$$

(85)  Dün     kitabı     Ahmet  bıraktı          masada.
      yesterday book+ACC Ahmet  leave+PAST+3SG table+LOC
      'It was Ahmet who left the book yesterday on the table.'

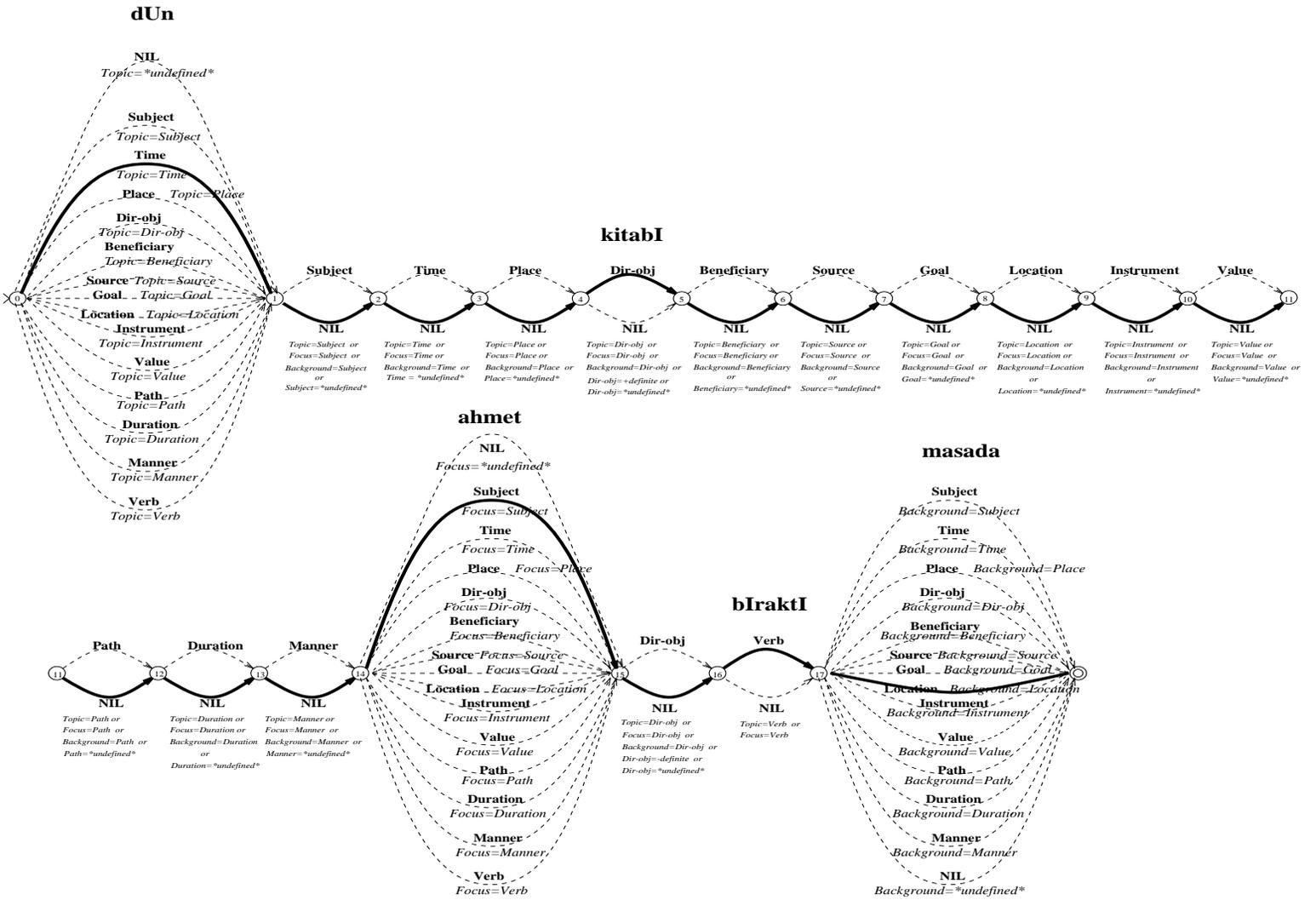

Figure 4.4: Transitions done to generate the example sentence.

**Existential Sentences**

We use the following general case-frame feature structure to encode the contents of an existential sentence:

$$\begin{bmatrix} \text{S-FORM} & \text{infinitive/adverbial/participle/finite} \\ \text{CLAUSE-TYPE} & \text{existential} \\ \text{VOICE} & \text{active/reflexive/reciprocal/passive/causative} \\ \text{SPEECH-ACT} & \text{imperative/optative/necessitative/wish/} \\ & \text{interrogative/declarative} \\ \text{QUES} & \begin{bmatrix} \text{TYPE} & \text{yes-no/wh} \\ \text{CONST} & \text{list-of(subject/dir-obj/etc.)} \end{bmatrix} \\ \text{VERB} & \begin{bmatrix} \text{ROOT} & \text{verb} \\ \text{POLARITY} & \text{negative/positive} \\ \text{TENSE} & \text{present/past/future} \\ \text{ASPECT} & \text{progressive/habitual/etc.} \\ \text{MODALITY} & \text{potentiality} \end{bmatrix} \\ \text{ARGUMENTS} & \begin{bmatrix} \text{SUBJECT} & \textit{c-name} \end{bmatrix} \\ \text{ADJUNCTS} & \begin{bmatrix} \text{TIME} & \textit{c-name} \\ \text{PLACE} & \textit{c-name} \end{bmatrix} \\ \text{CONTROL} & \begin{bmatrix} \text{IS} & \begin{bmatrix} \text{TOPIC} & \texttt{constituent} \\ \text{FOCUS} & \texttt{constituent} \\ \text{BACKGR} & \texttt{constituent} \end{bmatrix} \\ \text{ES} & \begin{bmatrix} \text{EVEN} & \texttt{constituent/poss-constituent} \\ \text{TOO} & \texttt{constituent/poss-constituent} \\ \text{QUES} & \texttt{constituent/poss-constituent} \end{bmatrix} \end{bmatrix} \end{bmatrix}$$

The value of the CONTROL | IS feature is again the information structure of the sentence. This case-frame differs from predicative sentences considering only arguments and adjuncts. Figure 4.5 gives an outline of our grammar for generating existential sentences.

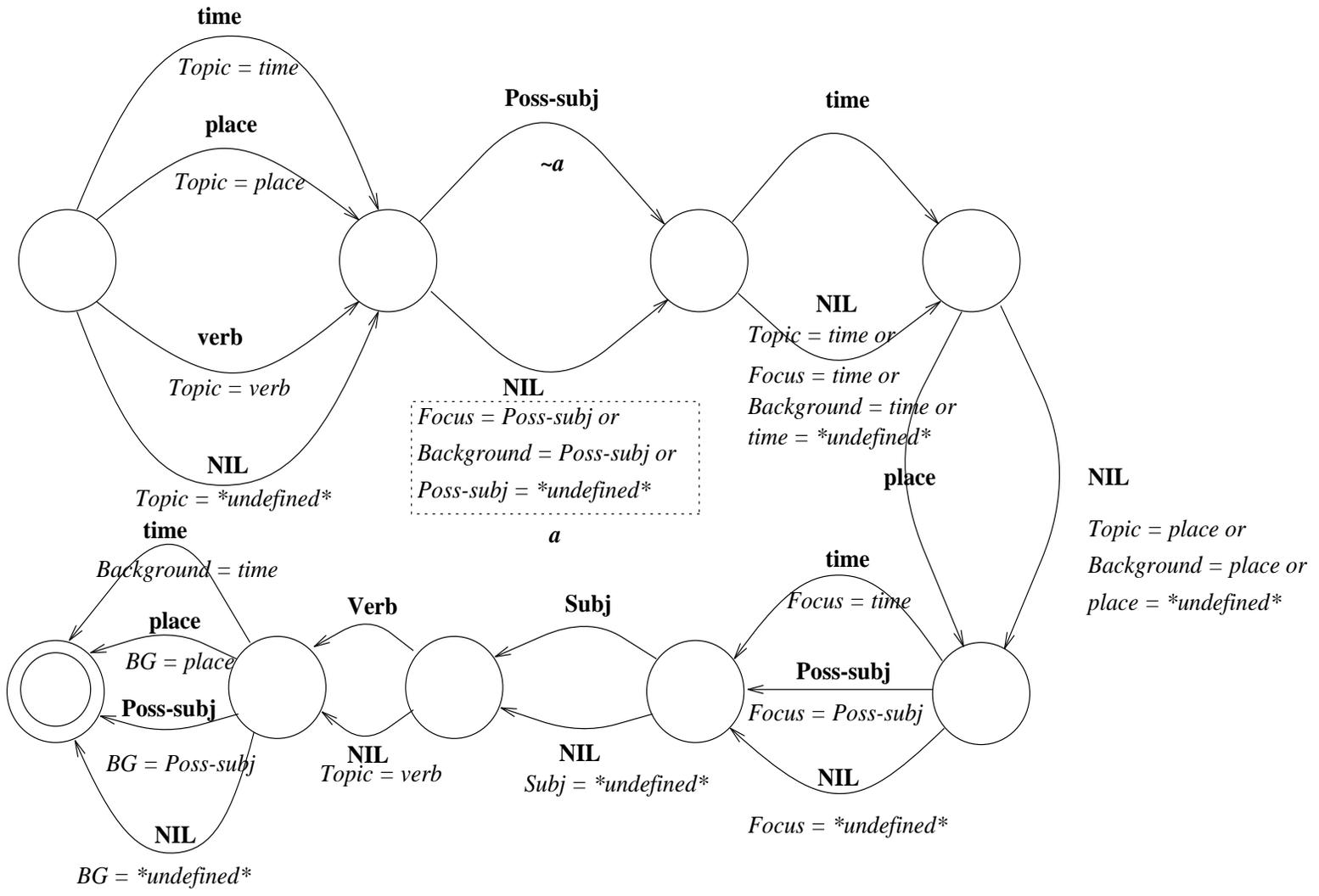

Figure 4.5: The finite state machine for existential sentences.

**Attributive Sentences**

We use the following general case-frame feature structure to encode the contents of an attributive sentence:

$$\begin{bmatrix}
\text{S-FORM} & \text{infinitive/adverbial/participle/finite} \\
\text{CLAUSE-TYPE} & \text{attributive} \\
\text{VOICE} & \text{active/reflexive/reciprocal/passive/causative} \\
\text{SPEECH-ACT} & \text{imperative/optative/necessitative/wish/} \\
& \text{interrogative/declarative} \\
\text{QUES} & \begin{bmatrix} \text{TYPE} & \text{yes-no/wh} \\ \text{CONST} & \text{list-of(subject/dir-obj/etc.)} \end{bmatrix} \\
\text{VERB} & \begin{bmatrix} \text{ROOT} & \text{verb} \\ \text{POLARITY} & \text{negative/positive} \\ \text{TENSE} & \text{present/past/future} \\ \text{ASPECT} & \text{progressive/habitual/etc.} \\ \text{MODALITY} & \text{potentiality} \end{bmatrix} \\
\text{ARGUMENTS} & \begin{bmatrix} \text{SUBJECT} & \text{c-name} \\ \text{PRED-PROPERTY} & \text{c-name} \end{bmatrix} \\
\text{ADJUNCTS} & \begin{bmatrix} \text{TIME} & \text{c-name} \\ \text{PLACE} & \text{c-name} \end{bmatrix} \\
\text{CONTROL} & \begin{bmatrix} \text{IS} & \begin{bmatrix} \text{TOPIC} & \text{constituent} \\ \text{FOCUS} & \text{constituent} \\ \text{BACKGR} & \text{constituent} \end{bmatrix} \\ \text{ES} & \begin{bmatrix} \text{EVEN} & \text{constituent/poss-constituent} \\ \text{TOO} & \text{constituent/poss-constituent} \\ \text{QUES} & \text{constituent/poss-constituent} \end{bmatrix} \end{bmatrix}
\end{bmatrix}$$

Figure 4.6 presents an outline of our grammar for generating attributive sentences. The CONTROL | IS feature in the case-frame, the information structure, is used to generate the appropriate surface order.

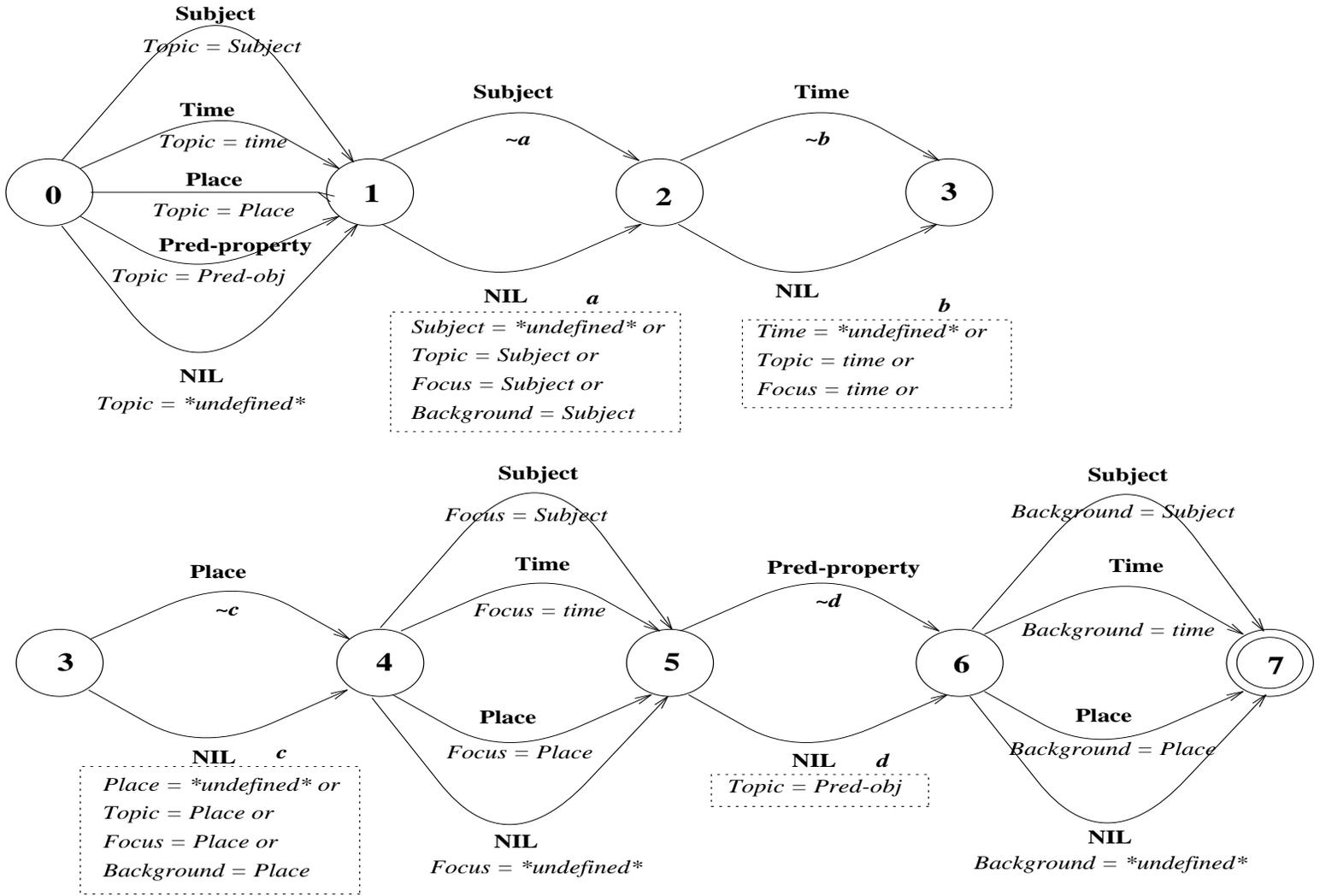

Figure 4.6: The finite state machine for attributive sentences.

## 4.1.3 Complex Sentences

Complex sentences are combinations of simple sentences (or complex sentences themselves) which are linked by either conjoining or various relationships like conditional dependence, cause–result, etc. The generator works on a feature structure representing a complex sentence which may be in one of the following forms:

- *a simple sentence.* In this case the sentence has the case-frame as its feature structure.

- *a series of simple or complex sentences* connected by coordinating or bracketing conjunctions. Such sentences have feature structures which have the individual case-frames as the values of their ELEMENTS features:

$$\begin{bmatrix} \text{TYPE} & \text{conj} \\ \text{CONJ} & \text{and/or/etc.} \\ \text{ELEMENTS} & \text{list-of(complex-sentence)} \end{bmatrix}$$

- *sentences linked with a certain relationship.* Such sentences have the feature structure:

$$\begin{bmatrix} \text{TYPE} & \text{linked} \\ \text{LINK-RELATION} & \text{rel} \\ \text{ARG1} & \text{complex-sentence} \\ \text{ARG2} & \text{complex-sentence} \end{bmatrix}$$

The case-frame of the following complex sentence, which is formed from two simple sentences linked by cause-result relationship, is given below it:

(86)  Sen   geldiğin           için      o    gitti.
      you   come+PART+P2SG  because   he   go+PAST+3SG
      '*He went because you came.*'

$$\begin{bmatrix} \text{TYPE} & \text{linked} \\ \text{LINK-RELATION} & \text{cause-result} \\ \text{ARG1} & \left\{ \begin{array}{l} \text{CASE-FRAME FOR} \\ \text{THE SENTENCE "Sen geldin."} \end{array} \right\} \\ \text{ARG2} & \left\{ \begin{array}{l} \text{CASE-FRAME FOR} \\ \text{THE SENTENCE "O gitti."} \end{array} \right\} \end{bmatrix}$$

### 4.1.4 Generating Noun Phrases

This subsection describes the generation of Turkish noun phrases and the feature structures (our input) that are used to denote them.

**The Design of a Semantic Representation for Turkish Noun Phrases**

**1. The Semantic Features for Common Nouns** The basic semantic information for concepts, needed in the generation of Turkish noun phrases are (with possible values are given in the parentheses):

      temporal (+/−),
      container (+/−),
      measure (+/−),
      countable (+/−),
      material (+/−).

These are needed only if the head is a common noun. These are available in the lexicon entries associated with the noun.

This semantic information is percolated from the head of the noun phrase to the whole noun phrase. Some of these features are not given in the input, but they can be obtained from the lexicon when needed. The usage of these features can be given as:

- The **temporal** feature is needed when determining the case of the noun phrase which expresses the location information (the specifying relation).

- The head of the noun phrase specifying a quantitative modifier of type *container-full* (explained in the following sections) should have the property of being a container, though this may be relaxed. The **container** feature is needed to ensure this, when necessary.

- In a similar way, the head of the noun phrase specifying a quantitative modifier of type *measure* (explained in the following sections) should have the property of being a unit, so **measure** feature is used to ensure this. This feature is also needed to ensure the absence of the word 'adet', 'tane', or 'parça' between a cardinal and a unit.

- The **countability** information is needed when comparing the countability of the head with that required by the quantifier.

- The **material** feature is used in a similar way with the container and measure features. The head of the noun phrase specifying a modifying relation (which is a noun phrase followed by the suffix `+DAn`) should have the property of being a material (although this may also be relaxed). It is needed to ensure this.

The generator needs the semantic information above for the whole noun phrase and information about its constituents, in order to generate the noun phrase. Besides these, it also needs emphasis information, to determine the word order. In the absence of any emphasis information, the generator generates the constituents in the default order.

In the following subsections, the feature structures for denoting noun phrases and their constituents are given.

**Proposed Feature Structures for the Noun Phrase**

We propose the following feature structure to describe the semantic structure of a noun phrase:

$$\begin{bmatrix} \text{REFERENT} & \begin{bmatrix} \text{ARG} & \left\{\begin{array}{l} \text{list-of}(\textit{c-name-no-spec-no-mod}) \\ \textit{b-con} \end{array}\right\} \\ \text{CONTROL} & [\text{DROP} \quad +/- \text{ (default }-)] \end{bmatrix} \\ \text{CLASSIFIER} & \textit{c-name-no-spec-qual} \\ \text{ROLES} & \textit{role-type} \\ \text{MODIFIER} & \begin{bmatrix} \text{MOD-REL} & \text{list-of}(\textit{m-type}) \\ \text{ORDINAL} & \begin{bmatrix} \text{ORDER} & \text{ilk/sonuncu/birinci/etc.} \\ \text{INTENSIFIER} & +/- \text{ (default }-) \end{bmatrix} \\ \text{QUAN-MOD} & \textit{q-type} \\ \text{QUALITATIVE} & \text{list-of}(\textit{s-prop}) \\ \text{CONTROL} & [\text{EMPHASIS} \quad \text{quantitative/qualitative}] \end{bmatrix} \\ \text{SPECIFIER} & \begin{bmatrix} \text{SET-SPEC} & \text{list-of}(\textit{c-name}) \\ \text{SPEC-REL} & \begin{bmatrix} \text{RELATION} & \text{dair/ait/location/etc.} \\ \text{ARGUMENT} & \text{list-of}(\textit{c-name}) \end{bmatrix} \\ \text{DEM} & \text{list-of}(\textit{d-type}) \\ \text{QUAN} & \begin{bmatrix} \text{QUANTIFIER} & \text{her/bazı/etc.} \\ \text{DEFINITE} & +/- \\ \text{REFERENTIAL} & +/- \\ \text{SPECIFIC} & +/- \end{bmatrix} \end{bmatrix} \\ \text{POSSESSOR} & \begin{bmatrix} \text{ARGUMENT} & \text{list-of}(\textit{c-name}) \\ \text{CONTROL} & \begin{bmatrix} \text{DROP} & +/- \text{ (default }-) \\ \text{MOVE} & +/- \text{ (default }-) \end{bmatrix} \end{bmatrix} \end{bmatrix}_{\textit{c-name}}$$

*c-name* is the type of the feature structure which denotes a noun phrase. *c-name* has a number of subtypes depending on the constraints on some of its features. The hierarchy of structures which are constrained forms of *c-name* is

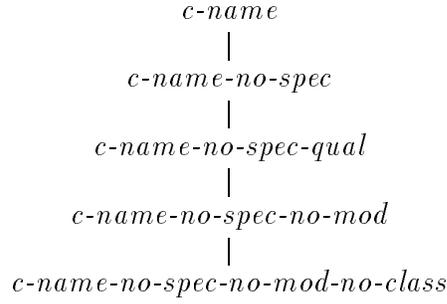

Figure 4.7: The hierarchy of subtypes of *c-name*

as in Figure 4.7. The subtypes of *c-name* are defined as follows:

1. If a noun phrase is not specified by any specifiers, then it has a structure of type *c-name-no-spec*. This structure has the following constraints on it:

$$c\text{-}name\text{-}no\text{-}spec \equiv \begin{bmatrix} \text{SPECIFIER} & \text{NIL} \\ \text{POSSESSOR} & \text{NIL} \end{bmatrix}_{c\text{-}name}$$

2. If a noun phrase is not specified by any specifiers, and furthermore it is not modified by any modifiers except the qualitative modifier, then it has a structure of type *c-name-no-spec-qual*. In this case, it has the constraints on it, which can be shown in the following feature structure:

$$c\text{-}name\text{-}no\text{-}spec\text{-}qual \equiv \begin{bmatrix} \text{MODIFIER} & \begin{bmatrix} \text{MOD-REL} & \text{NIL} \\ \text{ORDINAL} & \text{NIL} \end{bmatrix} \\ \text{SPECIFIER} & \text{NIL} \\ \text{POSSESSOR} & \text{NIL} \end{bmatrix}_{c\text{-}name}$$

3. If a noun phrase is not specified by any specifiers, and not modified by any modifiers, then it has a structure of type *c-name-no-spec-no-mod*. This structure has the constraints on it, which can be given with the following feature structure:

$$c\text{-}name\text{-}no\text{-}spec\text{-}no\text{-}mod \equiv \begin{bmatrix} \text{MODIFIER} & \text{NIL} \\ \text{SPECIFIER} & \text{NIL} \\ \text{POSSESSOR} & \text{NIL} \end{bmatrix}_{c\text{-}name}$$

CHAPTER 4. GENERATION OF TURKISH SENTENCES    664. If the structure of a noun phrase has only the REFERENT feature, and has a *b-con* (basic-concept, explained in detail in the following sections) structure as its value, then the type of its structure is *c-name-no-spec-no-mod-no-class*. The constraints on this structure can be given as below:

$$c\text{-}name\text{-}no\text{-}spec\text{-}no\text{-}mod\text{-}no\text{-}class \equiv \begin{bmatrix} \text{REFERENT} & \begin{bmatrix} \text{ARG} & b\text{-}con \end{bmatrix} \\ \text{CLASSIFIER} & \text{NIL} \\ \text{MODIFIER} & \text{NIL} \\ \text{SPECIFIER} & \text{NIL} \\ \text{POSSESSOR} & \text{NIL} \end{bmatrix}_{c\text{-}name}$$

The value of MODIFIER | CONTROL | EMPHASIS feature in a *c-name* structure is used to determine the order of quantitative and qualitative modifiers in the surface form. The value of this feature is closer to the head of the noun phrase in the surface form.

The value of a feature may be a list of structures, as explained in Chapter 3. This is shown as list-of(*str*) in the attribute-value matrix above, which is equivalent to the following feature structure:

$$\text{list-of}(str) \equiv \begin{bmatrix} \text{ELEMENTS} & \langle str, str, \ldots, str \rangle \\ \text{CONJ} & \text{and/or/none} \end{bmatrix}$$

The feature structures for the modifiers, specifiers and the classifier of the noun phrase are:

**1. Set Specifier**   The set specifier is also a noun phrase and the SET-SPEC feature has a structure of type *c-name* as value. Therefore, it has a structure like:

$$\begin{bmatrix} \text{SET-SPEC} & \text{list-of}(c\text{-}name) \end{bmatrix}$$

**2. Possessor**   The possessor is a list of the structures of type *c-name*. Sometimes, the possessor can be eliminated from the surface form as explained in Chapter 3, then some additional information regarding whether it can be omitted on the surface form may also be given. In a *c-name* structure, this information is given with the value of the POSSESSOR | CONTROL | DROP feature. If this feature has the value −, then the possessor can not be omitted, otherwise it is omitted. If no such information is given, then by default, this means that the possessor can not be omitted.

Furthermore, the possessor may change its position. The information about the movement of the possessor is given in a *c-name* structure with the POSSESSOR | CONTROL | MOVE feature. If this feature has + as value, then the possessor may move to a position behind the head in the surface form. So, MOVE can be used to encode possessor scrambling, when needed. The structure of possessor is given as the value of the feature POSS, and looks like:

$$\begin{bmatrix} \text{POSSESSOR} & \begin{bmatrix} \text{ARGUMENT} & \text{list-of}(\textit{c-name}) \\ \text{CONTROL} & \begin{bmatrix} \text{DROP} & +/- \text{ (default } -) \\ \text{MOVE} & +/- \text{ (default } -) \end{bmatrix} \end{bmatrix} \end{bmatrix}$$

**3. Specifying Relation** If specifying relation is present, it has the structure:

$$\begin{bmatrix} \text{SPEC-REL} & \begin{bmatrix} \text{RELATION} & \text{dair/ait/location} \\ \text{ARGUMENT} & \text{list-of}(\textit{c-name}) \end{bmatrix} \end{bmatrix}$$

The RELATION feature gives the relation of the argument noun phrase (the value of the ARGUMENT feature) with the head. If the relation is given by a postposition, then the RELATION feature has that postposition as its value. The RELATION feature has as value 'location', if the specifying relation mentions a temporal or spatial location. In such a case, the information whether the argument noun phrases should be in locative or nominative case is obtained from the value of SEM | TEMPORAL features of their referents.

**4. Demonstrative Specifier** The demonstrative specifier is a list of the Turkish demonstrative specifiers. If it is present, it has a structure like the following:

$$\left[\text{DEM} \quad \text{list-of}(d\text{-}type)\right]$$

where the structure of type *d-type* is the following:

$$_{d\text{-}type}\left[\text{DEMONS} \quad \text{bu/şu/o}\right]$$

**5. Quantifier** The quantifier is represented by the following structure:

$$\left[\text{QUAN} \quad \begin{bmatrix} \text{QUANTIFIER} & \text{her/bazı/etc.} \\ \text{DEFINITE} & +/- \\ \text{REFERENTIAL} & +/- \\ \text{SPECIFIC} & +/- \end{bmatrix}\right]$$

The QUANTIFIER feature has as value the root of the quantifier. If the quantifier feature has NIL as value, then the DEFINITE, REFERENTIAL, and SPECIFIC features determine the presence of the indefinite article 'bir' (a/an). Furthermore, these features, together with the role of the noun phrase, determine the case and the position of noun phrase in the sentence.

**6. Modifying Relation** A noun phrase may have more than one modifying relations, so the MOD-REL feature of a concept has as value a list of modifying relations. The feature structures which are the elements of this list are of *m-type*, which has the following structure:

$$_{m\text{-}type}\left[\begin{matrix} \text{RELATION} & \text{gibi/kadar/made-of/with/etc.} \\ \text{ARGUMENT} & \text{list-of}(c\text{-}name) \end{matrix}\right]$$

If an element of the modifying relation list is a postpositional phrase, where the postposition gives the relation of some other concepts with the head, the RELATION feature has that postposition as its value. If an element of the modifying relation list is an adjectival phrase made from a noun phrase by the suffix:

- **+DAn**, then the value of RELATION feature is 'made-of'. The argument noun

phrases must have the property of being a material, but this may also be relaxed.

- **+lH**, then the value of RELATION feature is 'with'. The argument has on it the constraint that it should be a list of structures of type *c-name-no-spec*.

- **+sHz**, then the value of RELATION feature is 'without'. The argument has on it the constraint that it should be a list of structures of type *c-name-no-spec-qual*.

- **+DA**, then the value of RELATION feature is 'made-on'. The argument has on it the constraint that it should be a list of structures of type *c-name-no-spec-qual*.

- **+lHk**, then the value of RELATION feature is 'of'. However, in this case the value of the ARGUMENT feature is not a list of structures of noun phrases, but it is the structure of a quantitative modifier of type *measure* (the structure of a quantitative modifier of type *measure* will be explained when explaining quantitative modifiers). But the value of the APPROX feature in this structure should be –, because the quantitative modifier should not have the word 'civarında' as the head.

**7. Ordinal** The ordinal also has a structure similar to the demonstrative specifier or quantifier. When it is present, it has the structure:

$$\begin{bmatrix} \text{ORDINAL} & \begin{bmatrix} \text{ORDER} & \text{ilk/sonuncu/birinci/etc.} \\ \text{INTENSIFIER} & +/- \text{ (default -)} \end{bmatrix} \end{bmatrix}$$

**8. Quantitative Modifier** The QUAN-MOD feature contains the quantitative modifier. In Turkish, quantity information can be expressed in several forms, as described in Chapter 3. The type of the feature structure (*q-type*) which gives the quantity information can be one of *number, measure, container-full* or *fuzzy-quantity*, with the following structures:

1. If the quantity is expressed by a cardinal alone, as in the following example,

(87)     dört elma
         `four apple`
         *'four apples'*

then the feature structure corresponding to the quantitative modifier of this noun phrase will be as follows:

$$\begin{bmatrix} \text{LOW} & 4 \\ \text{HIGH} & \text{NIL} \\ \text{CONTROL} & \begin{bmatrix} \text{FORMAL-CARD} & - \text{(default } -) \\ \text{FORMAL-LOW} & - \text{(default } -) \\ \text{FORMAL-HIGH} & - \text{(default } -) \end{bmatrix} \end{bmatrix}_{number}$$

Here, the value of the QUAN-MOD | CONTROL | FORMAL-CARD feature determines whether the cardinal should be followed by one of the words: 'adet', 'tane', or 'parça'. In the absence of this information, the cardinal is not followed by any of these words. The values of the features QUAN-MOD | CONTROL | FORMAL-LOW and QUAN-MOD | CONTROL | FORMAL-HIGH determine the presence of the words 'en az' (at least) and 'en çok' (at most), respectively. These words are used in order to specify limits.

2. If the quantity is expressed by a range of cardinals, as in the following example,

(88)     iki üç elma
         `two three apple`
         *'2 to 3 apples'*

then the feature structure corresponding to the quantitative modifier of this noun phrase will be like:

$$\text{number}\begin{bmatrix} \text{LOW} & 2 \\ \text{HIGH} & 3 \\ \text{CONTROL} & \begin{bmatrix} \text{FORMAL-CARD} & - \text{ (default } -) \\ \text{FORMAL-LOW} & - \text{ (default } -) \\ \text{FORMAL-HIGH} & - \text{ (default } -) \end{bmatrix} \end{bmatrix}$$

Here the roles of QUAN-MOD | CONTROL | FORMAL-CARD, QUAN-MOD | CONTROL | FORMAL-LOW and QUAN-MOD | CONTROL | FORMAL-HIGH features are the same as in the case of cardinal. If the value of QUAN-MOD | CONTROL | FORMAL-LOW feature was + then (89a) would be generated from the above feature structure, whereas if the values of QUAN-MOD | CONTROL | FORMAL-LOW and QUAN-MOD | CONTROL | FORMAL-HIGH features were both +, then (89b) would be generated from it:

(89) a. en    az    iki   üç
    `most less two three`
    'at least two to three'

   b. en    az    iki   en    çok   üç
    `most less two most much three`
    'at least two, at most three'

3. If the quantity is expressed by a noun phrase whose head is a measure noun as in example (53c) of Chapter 3, then the corresponding feature structure will be:

$$\text{measure}\begin{bmatrix} \text{QUANTITY} & \text{number} \\ \text{UNIT} & \text{c-name-no-spec-no-mod-no-class} \\ \text{APPROX} & +/- \text{ (default } -) \end{bmatrix}$$

The value of the QUAN-MOD | APPROX feature, which is by default − in the absence of this feature, determines whether the word 'civarında' (meaning 'about') should be present.

So, the MODIFIER | QUAN-MOD feature of the noun phrase

(90) a. bir iki   kilo   elma
   one two kilo apple
   '1 to 2 kilos of apples'

will have as its value, the following feature structure:

$$\begin{bmatrix} \text{QUANTITY} & number\begin{bmatrix} \text{LOW} & 1 \\ \text{HIGH} & 2 \end{bmatrix} \\ \text{UNIT} & \textit{c-name}\begin{bmatrix} \text{REFERENT} & \textit{b-con}\begin{bmatrix} \text{ARG} & \begin{bmatrix} \text{CONCEPT} & \#\text{kilo} \\ \text{SEM} & \begin{bmatrix} \text{MEASURE} & + \end{bmatrix} \end{bmatrix} \end{bmatrix} \end{bmatrix} \end{bmatrix}_{measure}$$

4. If the quantity is expressed by a noun phrase, whose head is a container, as in example (53a) of Chapter 3, the feature structure which is the value of the quantity feature, will be similar to the above two. The only difference will be the type of the feature structure, this time it will be *container-full* instead of *measure*, and there will not be a QUAN-MOD | APPROX feature. So, the corresponding feature structure will be as follows:

$$\textit{container-full}\begin{bmatrix} \text{QUANTITY} & number \\ \text{UNIT} & \textit{c-name-no-spec-qual} \end{bmatrix}$$

5. If the quantity is expressed by a fuzzy quantity, as in example (52) of Chapter 3, then the corresponding part of the feature structure of such a noun phrase will be like:

$$\textit{fuzzy-quantity}\begin{bmatrix} \text{F-QUAN} & \text{çok} \\ \text{CONTROL} & \begin{bmatrix} \text{FORMAL-QUANTITY} & +/- \text{ (default } -) \end{bmatrix} \end{bmatrix}$$

Some of these adjectives may also be followed by one of the words 'miktarda' (with uncountable heads) or 'sayıda' (with countable heads). The value of the feature CONTROL | FORMAL-QUANTITY is used when determining the

presence of the word 'miktarda' or 'sayıda'. If it is +, then one of these words should be present.

**9. Qualitative Modifier**  The head of a noun phrase may be modified by more than one qualitative modifiers. Therefore the QUALITATIVE feature of a concept is a list of individual qualitative modifiers. The elements of this list are of type simple property (i.e. adjective), which has the following structure:

$$s\text{-}prop\begin{bmatrix} \text{P-NAME} & \text{basic-property} \\ \text{INTENSIFIER} & i\text{-}type \end{bmatrix}$$

In order to be modified by an intensifier, the basic-property should be gradable (i.e. it should have the value + for its GRADABLE feature). The intensifiers of the simple properties can be adverbs mentioning the degree of that property, as in example (59a) (Chapter 3), or they can be postpositional phrases comparing that property with another, as in example (59b) (Chapter 3). Therefore, the type of the structure for the intensifier ($i$-$type$) can be: *degree* or *p-comparative*. These two have the following structures, respectively:

$$degree\begin{bmatrix} \text{DEGREE} & \text{çok/en/daha az/etc.} \end{bmatrix}$$

$$p\text{-}comparative\begin{bmatrix} \text{COMPARATOR} & \text{daha/kadar} \\ \text{ARG} & c\text{-}name \end{bmatrix}$$

The intensifier of type *degree* can be 'açık' (light) or 'koyu' (dark), if it is modifying a color (i.e. the basic-property has the value + for its COLOR feature).

**10. The Classifier and the Head**  The head of a noun phrase is given as the value of the REFERENT feature, and the classifier is given as the value of the CLASSIFIER feature, which has the structure of *c-name* with certain constraints on it. Therefore, the REFERENT and CLASSIFIER features have the following structures:

$$\left[\begin{array}{ll} \text{REFERENT} & \left[\begin{array}{ll} \text{ARG} & \left\{\begin{array}{l} \text{list-of}(\textit{c-name-no-spec-no-mod}) \\ \textit{b-con} \end{array}\right\} \\ \text{CONTROL} & \left[\text{DROP} \quad +/- \text{ (default } -)\right] \end{array}\right] \\ \text{CLASSIFIER} \quad \textit{c-name-no-spec-qual} \end{array}\right]_{\textit{c-name}}$$

The referent may either have a list of *c-name-no-spec-no-mod* structures or a *b-con* structure. The structure of a typical *b-con* is:

$$\left[\begin{array}{ll} \text{CONCEPT} & \#\text{köpek} \\ \text{SEM} & \left[\begin{array}{ll} \text{TEMPORAL} & +/- \\ \text{CONTAINER} & +/- \\ \text{MEASURE} & +/- \\ \text{COUNTABLE} & +/- \\ \text{MATERIAL} & +/- \end{array}\right] \end{array}\right]_{\textit{b-con}}$$

The value of the REFERENT | CONTROL | DROP feature determines whether the head will drop or not. If this feature has a value +, then the head will drop, as explained in Chapter 3. This feature has a value − by default.

The following is an example to a complex feature structure:

$$\left[\begin{array}{ll} \text{REFERENT} & \left[\begin{array}{ll} \text{ARG} & \left[\begin{array}{ll} \text{REFERENT} & \left[\text{ARG}_{\textit{b-con}}\left[\text{CONCEPT} \quad \#\text{oran}\right]\right] \\ \text{CLASSIFIER} & \left[\text{REFERENT} \left\{\begin{array}{l} \text{ARG OF TYPE } \textit{b-con} \\ \text{WITH CONCEPT } \#\text{komisyon} \end{array}\right\}\right]_{\textit{c-name}} \end{array}\right]_{\textit{c-name}} \end{array}\right] \\ \text{CLASSIFIER} & \left[\begin{array}{ll} \text{REFERENT} & \left[\text{ARG}_{\textit{b-con}}\left[\text{CONCEPT} \quad \#\text{kart}\right]\right] \\ \text{CLASSIFIER} & \left[\text{REFERENT} \left[\text{ARG}_{\textit{b-con}}\left[\text{CONCEPT} \quad \#\text{kredi}\right]\right]\right]_{\textit{c-name}} \end{array}\right]_{\textit{c-name}} \end{array}\right]_{\textit{c-name}}$$

which is the feature structure of the noun phrase:

(91) a. kredi   kartı      komisyon    oranı
       credit card+P3SG commission rate+P3SG
       'credit card commission rate'

and the feature structure of the noun phrase:

(92) a. komisyon    oranı
       commission rate+P3SG
       'commission rate'

can be given with the following feature structure:

$$\begin{bmatrix} \text{REFERENT} & \begin{bmatrix} \text{ARG}_{b\text{-}con} \begin{bmatrix} \text{CONCEPT} & \#\text{oran} \end{bmatrix} \end{bmatrix} \\ \text{CLASSIFIER} & \begin{bmatrix} \text{REFERENT} & \begin{bmatrix} \text{ARG}_{b\text{-}con} \begin{bmatrix} \text{CONCEPT} & \#\text{komisyon} \end{bmatrix} \end{bmatrix} \end{bmatrix}_{c\text{-}name} \end{bmatrix}_{c\text{-}name}$$

**Roles**

The sentential modifiers of the noun phrase and the noun phrases which are sentential clauses are linked to our feature structure denoting noun phrases, by the ROLES feature.

The sentential modifiers of the head are gapped sentential clauses, where the head is the filler of the gap. Therefore, if the head has a sentential modifier, the ROLES feature in the feature structure for this noun phrase has as value a feature structure which can be given as follows:

$$\begin{bmatrix} \text{ROLE} & \text{agent/patient/theme/etc.} \\ \text{ARG} & \textit{case-frame} \end{bmatrix}$$

Here, the ROLE feature has as value the role of the head, the gapped constituent in the gapped sentential clause and the ARG feature has the case-frame for the sentential clause. The case-frames for sentential clauses are very similar to the case-frames for sentences.

The noun phrases which are full sentential clauses are acts, facts and adverbials. If the noun phrase is a full sentential clause, then the feature structure denoting it has only the ROLES feature with the following feature structure as value:

$$\begin{bmatrix} \text{TYPE} & \text{ind-act/def-act/fact/etc.} \\ \text{ARG} & \textit{case-frame} \end{bmatrix}$$

The TYPE feature gives the type of the sentential clause and the ARG feature contains the case-frame for the sentential clause.

The form of the verb of the sentential clause is determined using the ROLE feature with sentential modifiers and TYPE feature with full sentential clauses which act as noun phrases.

### Some Example Feature Structures

The following are some examples for the feature structure of a noun phrase:

- yedi    adamdan   ikisi
  seven man+LOC two+P3SG
  *'two of the seven men'*

$$\begin{bmatrix} \text{REFERENT} & \begin{bmatrix} \text{ARG} & {}_{b\text{-}con}\begin{bmatrix} \text{CONCEPT} & \#\text{adam} \end{bmatrix} \\ \text{CONTROL} & \begin{bmatrix} \text{DROP} & + \end{bmatrix} \end{bmatrix} \\ \text{MODIFIER} & \begin{bmatrix} \text{QUAN-MOD} & {}_{number}\begin{bmatrix} \text{LOW} & 2 \\ \text{HIGH} & \text{NIL} \end{bmatrix} \end{bmatrix} \\ \text{SPECIFIER} & \begin{bmatrix} \text{SET-SPEC} & \begin{bmatrix} \text{REFERENT} & \begin{bmatrix} \text{ARG} & {}_{b\text{-}con}\begin{bmatrix} \text{CONCEPT} & \#\text{adam} \end{bmatrix} \end{bmatrix} \\ \text{MODIFIER} & \begin{bmatrix} \text{QUAN-MOD} & {}_{number}\begin{bmatrix} \text{LOW} & 7 \\ \text{HIGH} & \text{NIL} \end{bmatrix} \end{bmatrix} \end{bmatrix}_{c\text{-}name} \end{bmatrix}_{c\text{-}name}$$

- at kadar büyük bir köpek
  horse as big a dog
  *'a dog as big as a horse'*

$$\begin{bmatrix} \text{REFERENT} & \begin{bmatrix} \text{ARG} & \begin{bmatrix} \text{CONCEPT} & \#\text{köpek} \end{bmatrix}_{b\text{-}con} \end{bmatrix} \\ \text{MODIFIER} & \begin{bmatrix} \text{QUALITATIVE} & \begin{bmatrix} \text{P-NAME} & \#\text{büyük} \\ \text{INTENSIFIER} & \begin{bmatrix} \text{COMPARATOR} & \text{kadar} \\ \text{ARG} & \begin{Bmatrix} \text{A } c\text{-}name \text{ STRUCTURE} \\ \text{WITH A REFERENT OF} \\ \text{TYPE } b\text{-}con \text{ WITH} \\ \text{CONCEPT } \#\text{at} \end{Bmatrix} \end{bmatrix}_{p\text{-}comparative} \end{bmatrix}_{s\text{-}prop} \end{bmatrix} \\ \text{SPECIFIER} & \begin{bmatrix} \text{QUAN} & \begin{bmatrix} \text{QUANTIFIER} & \text{NIL} \\ \text{DEFINITE} & - \\ \text{REFERENTIAL} & + \\ \text{SPECIFIC} & - \end{bmatrix} \end{bmatrix} \end{bmatrix}_{c\text{-}name}$$

**Problems in Generating Noun Phrases**

The surface form of a noun phrase with the following feature structure:

$$\begin{bmatrix} \text{REFERENT} & \begin{bmatrix} \text{ARG} & \begin{bmatrix} \text{CONCEPT} & \#\text{adam} \end{bmatrix}_{b\text{-}con} \end{bmatrix} \\ \text{MODIFIER} & \begin{bmatrix} \text{QUAN-MOD} & \begin{bmatrix} \text{LOW} & 2 \\ \text{HIGH} & \text{NIL} \end{bmatrix}_{number} \\ \text{MOD-REL} & \begin{bmatrix} \text{ARGUMENT} & \begin{bmatrix} \text{REFERENT} & \begin{Bmatrix} \text{ARGUMENT OF TYPE } b\text{-}con \\ \text{WITH CONCEPT } \#\text{kalem} \end{Bmatrix} \end{bmatrix}_{c\text{-}name} \end{bmatrix}_{with} \end{bmatrix} \end{bmatrix}_{c\text{-}name}$$

can be both of the below phrases, depending on the emphasis information:

(95) a. iki, kalemli adam
   two pencil+WITH man
   *'two men with a pencil'*

b. kalemli    iki  adam
   `pencil+WITH two man`
   *'two men with a pencil'*

But if no comma is generated in the first surface form, it will be ambiguous, because it is the surface form of the following feature structure also:

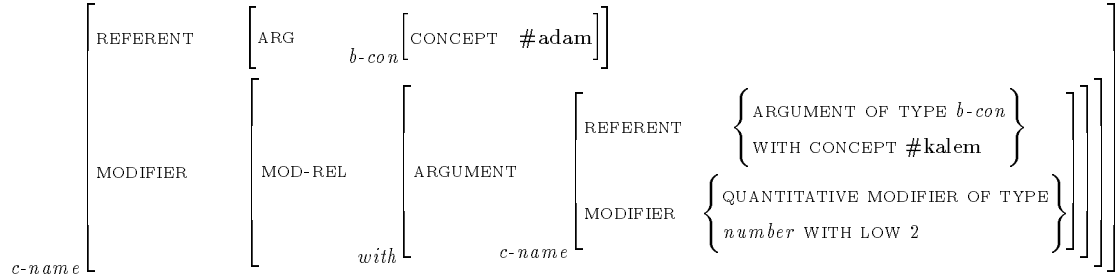

Therefore, if the generator receives no emphasis information, it always generates the second surface form, for the first feature structure, because generating a comma between the modifying relation and the quantitative modifier, in order to eliminate this ambiguity, is not very good stylistically.

A similar ambiguity is also present in the surface form corresponding to the following feature structure. This phrase has a qualitative modifier and a modifying relation:

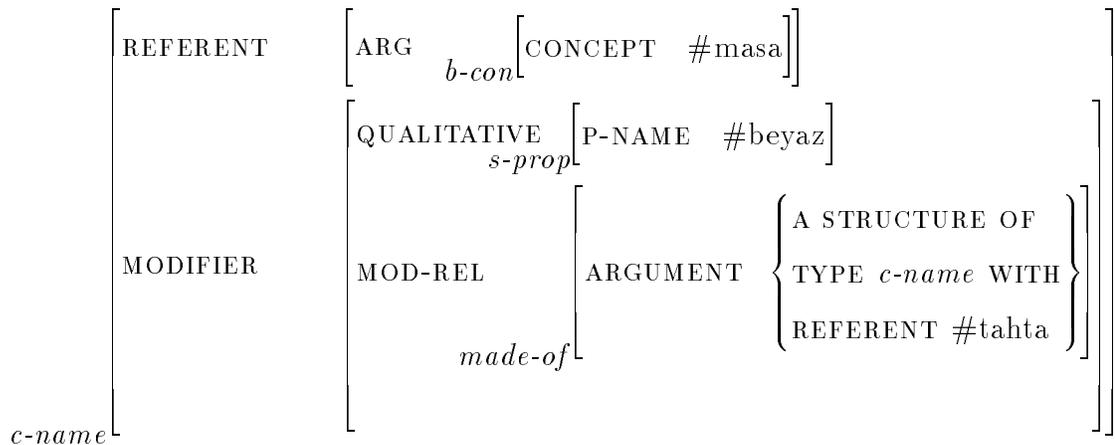

And the corresponding surface forms are:

(96) a. beyaz, tahtadan  masa
      white wood+ABL table
      'white table made of wood'

   b. tahtadan  beyaz  masa
      wood+ABL white table
      'white table made of wood'

The first of these surface forms can also be ambiguous if the comma is not present in the surface form, because it also corresponds to the following feature structure:

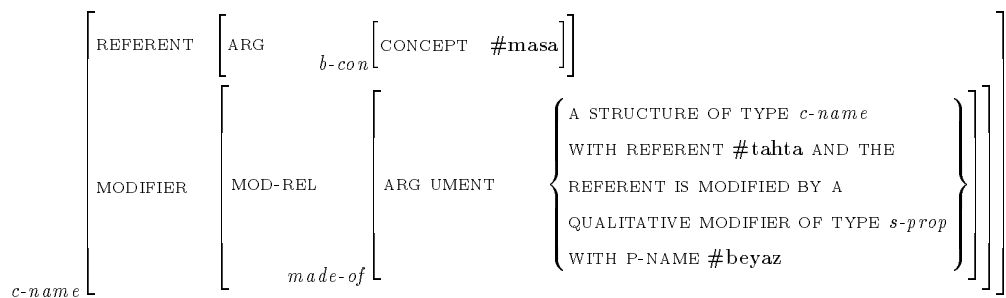

The generator should generate the surface forms that are not ambiguous. In order to realize this,

- If there is no emphasis information, the set of modifying relations precede the quantitative modifier in the surface form,

- If there is emphasis information, there will be a comma between each element of the set of modifying relations and between the modifying relations and quantitative modifier.

## 4.2   Grammar Architecture

Our generation grammar is written in a formalism called Pseudo Unification Grammar implemented by the GenKit generation system [22], developed at Carnegie Mellon University–Center for Machine Translation. In the following subsections, information about GenKit, and then some example rules of our grammar will be given.

## 4.2.1 GenKit

Generation Kit (GenKit) is a system which compiles a grammar into a sentence generation program. The grammar formalism used by GenKit is also called *Pseudo Unification Grammar* [22]. Each rule of the grammar consists of a context-free phrase structure description and a set of *feature constraint equations*, which are used to express constraints on feature values. Non-terminals in the phrase structure part of a rule are referenced as x0,...,xn in the equations, where x0 corresponds to the non-terminal in the left hand side, and xn is the $n^{th}$ non-terminal in the right hand side.

To implement the sentence level generator (described by the finite state machine presented earlier), we use rules of the form:

$$S_i \rightarrow XP\ S_j$$

where the $S_i$ and $S_j$ denote some state in the finite state machine and the XP denotes the constituent to be realized while taking the transition between states $S_i$ and $S_j$, labeled XP. By the feature constraint equations, the corresponding part of the feature structure which was assigned to $S_i$ previously, is assigned to XP, and the remaining is assigned to $S_j$. If this XP corresponds to a sentential clause, the same set of rules are recursively applied. This is a variation of the method suggested by Takeda *et al.* [21]. By this kind of rules, there is no need to write a separate rule for each possible constituent order.

In generation, non-determinism, producing multiple surface forms for a given input, is a serious problem. If there is no style-related information (an information structure) in the input, our generator generates the sentence in a default order. If there was no such a default order, non-determinism would have been a big problem, because Turkish is a free-constituent order language. For a given input of two arguments plus the verb (without no information structure) 6 surface sentences (all combinations) might be generated.

Since the context-free rules are directly compiled into tables, the performance of the system is essentially independent of the number of rules, but depends on the complexity of the feature constraint equations (which are compiled into LISP code). Currently, our grammar has 273 rules (excluding lexical rules), each with very simple constraint checks. Of these 273 rules, 133 are for sentences and 107 are for noun phrases, and the remaining are for adverbials and verbs.

### 4.2.2 Example Rules

Figure 4.8 demonstrates the transitions from state 0 to state 1 of our finite state machine for generating predicative sentences. The following are the simplified

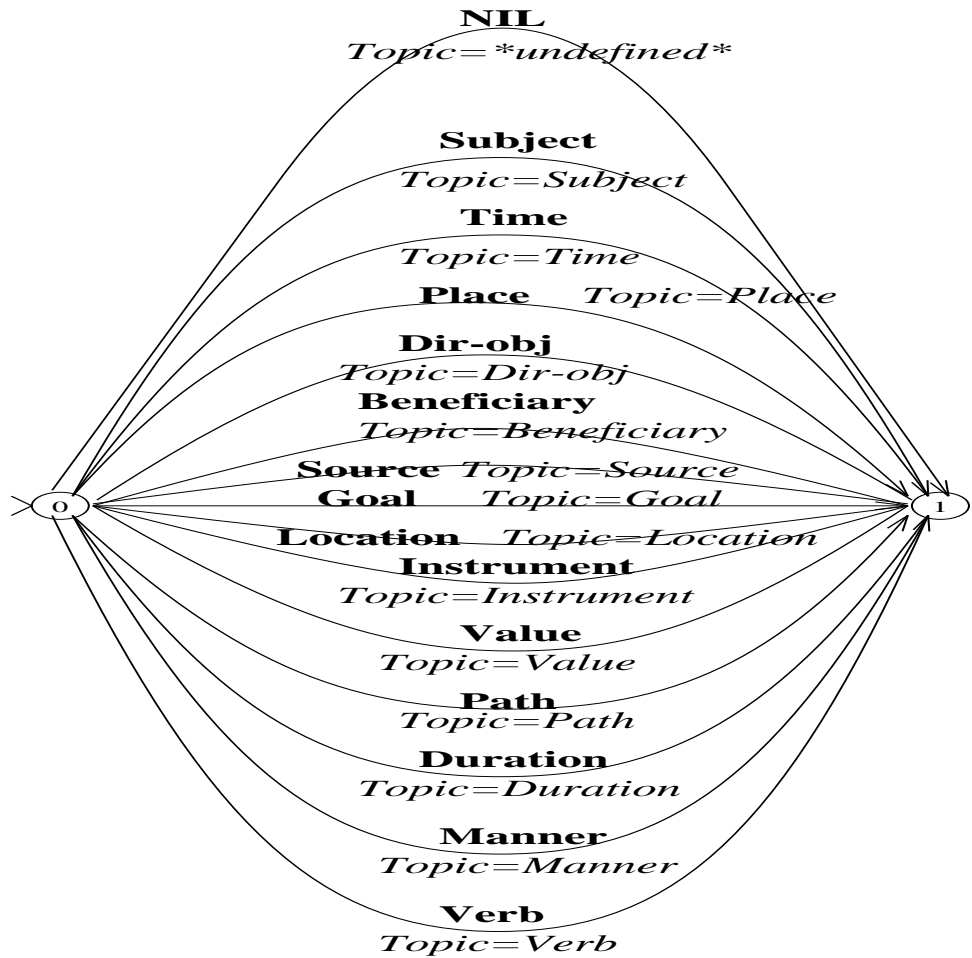

Figure 4.8: Transitions from state 0 to state 1 of our finite state machine for generating predicative sentences.

forms of our rules for the top three transitions in Figure 4.8:

```
(<S> <==> (<S1>)
    (
     ((x0 control topic) =c *undefined*)
     (x1 = x0))
    )

(<S> <==> (<Subject> <S1>)
    (
     ((x0 control topic) =c subject)
     (x2 = x0)
     ((x2 arguments subject) = *remove*)
     (x1 = (x0 arguments subject)))
    )

(<S> <==> (<Time> <S1>)
    (
     ((x0 control topic) =c time)
     (x2 = x0)
     ((x2 adjuncts time) = *remove*)
     (x1 = (x0 adjuncts time)))
    )
```

The first rule above is for the **NIL** transition, this transition is done if the topic is not defined. The second rule is for the transition labeled **Subject**, if the topic is subject, then this transition is done. In the feature constraint equations, it is checked whether the subject is the topic, and if so, the part of the feature structure for subject is assigned to <Subject>, and the remaining is assigned to <S1>. The third rule is for the transition labeled **Time**.

The grammar also has rules for realizing a constituent like <Subject> or <Time> (which may eventually call the same rules if the argument is sentential) and rules like above for traversing the finite state machine from state 1 to the final state.

## 4.3  Interfacing with Morphology

As Turkish has complex agglutinative word forms with productive inflectional and derivational morphological processes, we handle morphology outside our system using the generation component of a full-scale morphological analyzer of Turkish [17]. Within GenKit, we generate relevant abstract morphological features such as:

- agreement,
- possessive, and
- case

markers for nominals and

- voice,
- polarity,
- tense,
- aspect,
- mood, and
- agreement

markers for verbal forms. This information is properly ordered at the interface and sent to the morphological generator, which then:

1. performs concrete morpheme selection, dictated by the morphotactic constraints and morphophonological context,

2. handles morphographemic phenomena such as vowel harmony, and vowel and consonant ellipsis, and

3. produces an agglutinative surface form.

For example, the following feature structures are the outputs of our generator for nominal and verbal forms, respectively. These are sent to the morphological generator, which then performs morpheme selections and converts them into the intermediate forms below, and produces the agglutinative surface forms from these intermediate forms:

$$\begin{bmatrix} \text{CAT} & \text{NOUN} \\ \text{ROOT} & \text{kalem} \\ \text{AGR} & \text{3SG} \\ \text{POSS} & \text{1SG} \\ \text{CASE} & \text{GEN} \end{bmatrix}$$

$$\downarrow$$

$$\text{kalem}+\emptyset+\text{Hm}+\text{Hn}$$

$$\downarrow$$

$$\text{kalemimin}$$

$$\begin{bmatrix} \text{CAT} & \text{VERB} \\ \text{ROOT} & \text{gel} \\ \text{SENSE} & \text{POS} \\ \text{TAM1} & \text{PROG1} \\ \text{AGR} & \text{1SG} \end{bmatrix}$$

$$\downarrow$$

$$\text{gel}+\emptyset+\text{Hyor}+\text{Hm}$$

$$\downarrow$$

$$\text{geliyorum}$$

## 4.4  Comparison With Related Work

There were two studies done on generation of Turkish sentences, previously. The first one is the M.Sc. thesis of Colin Dick [3], done in the Department of Artificial Intelligence, University of Edinburgh, and the second one is the Ph.D. thesis of Beryl Hoffman, at the Computer and Information Science Department of University of Pennsylvania.

Dick [3] has worked on a classification based language generator for Turkish. His goal was to generate Turkish sentences of varying complexity, from input semantic representations in Penman's Sentence Planning Language (SPL). However, his generator is not complete, in that, noun phrase structures in their entirety, postpositional phrases, word order variations, and many morphological phenomena are not implemented. Our generator differs from his in various aspects: We use a case-frame based input representation which we feel is more suitable for languages with free constituent order. Our coverage of the grammar is substantially higher than the coverage presented in his thesis and we also use a full-scale external morphological generator to deal with complex morphological phenomena of agglutinative lexical forms of Turkish, which he has attempted embedding into the sentence generator itself.

Hoffman, in her thesis [8, 9], has used the Multiset–Combinatory Categorial Grammar formalism [10], an extension of Combinatory Categorial Grammar to handle free word order languages, to develop a generator for Turkish. Her generator also uses relevant features of the information structure of the input and can handle word order variations within embedded clauses. She can also deal with scrambling out of a clause dictated by information structure constraints, as her formalism allows this in a very convenient manner. The word order information is lexically kept as multisets associated with each verb. She has demonstrated the capabilities of her system as a component of a prototype database query system. We have been influenced by her approach to incorporate information structure in generation, but, since our aim is to build a wide-coverage generator for Turkish for use in a machine translation application, we have opted to use a simpler formalism and a very robust implementation environment. Our generator also differs from her generator, in that: When the information structure for sentences (explained in detail in Chapter 3 of this thesis) is not present, we generate sentences in a default order (which is also given in Chapter 3), whereas her generator generates nothing in such a case.

A concurrent study was also done by Turgay Korkmaz [14]. He used a functional linguistic theory called Systemic-Functional grammar, and the FUF text generation system to implement a sentence generator for Turkish. His generator takes semantic descriptions of sentences and then produces a morphological description for each lexical constituent of the sentence. His generator does not handle long-distance scramblings, unbounded dependencies and discontinuous constituents, but is the first one to use the systemic functional approach in the context of Turkish.

# Chapter 5

# Example Outputs and Future Work

In this chapter, we present some example outputs of our generator. In Appendix C, some additional examples from computer manuals are also given.

## 5.1 Example Outputs

In this section, some examples to demonstrate the work of our generator are given.

**Example 1:**

Input:

```
; Adam elmayI kadIna verdi
; This sentence is in the default order, so there will not be
; a CONTROL feature.

((s-form finite)
 (clause-type predicative)
 (voice active)
```

```
      (speech-act declarative)
      (verb
       ((root "ver")
        (sense positive)
        (tense past)
        (aspect perfect)))
      (arguments
       ((subject
          ((referent
             ((arg
                ((concept "adam")))
              (agr ((number singular)
                    (person 3))))))))
         (dir-obj
          ((referent
             ((arg
                ((concept "elma")))
              (agr
                ((number singular)
                 (person 3)))))
           (specifier
            ((quan
              ((definite +)))))))
         (goal
          ((referent
             ((arg
                ((concept "kadIn")))
              (agr
                ((number singular)
                 (person 3)))))))))))

Output:

  [[CAT=NOUN][ROOT=adam][AGR=3SG][POSS=NONE][CASE=NOM]] --> adam
  [[CAT=NOUN][ROOT=elma][AGR=3SG][POSS=NONE][CASE=ACC]] --> elmayI
```

```
[[CAT=NOUN][ROOT=kadIn][AGR=3SG][POSS=NONE][CASE=DAT]] --> kadIna
[[CAT=VERB][ROOT=ver][SENSE=POS][TAM1=PAST][AGR=3SG]] --> verdi
[PERIOD] --> .
```

**Example 2:**

```
Input:

  ; KadIna adam verdi elmayI
  ; This sentence is not in the default order, the destination/recipient,
  ; ``kadIn'', is the topic, the subject, ``adam'', is the focus,
  ; the direct object, ``elma'', is the background.

  ((s-form finite)
   (clause-type predicative)
   (voice active)
   (speech-act declarative)
   (verb
    ((root "ver")
     (sense positive)
     (tense past)
     (aspect perfect)))
   (arguments
    ((subject
       ((referent
          ((arg
             ((concept "adam")))
            (agr ((number singular)
                  (person 3))))))))
      (dir-obj
       ((referent
          ((arg
             ((concept "elma")))
            (agr
             ((number singular)
              (person 3)))))
         (specifier
          ((quan
             ((definite +))))))))
```

```
       (goal
        ((referent
           ((arg
              ((concept "kadIn")))
             (agr
              ((number singular)
               (person 3))))))))
    (control
     ((is
        ((topic goal)
         (focus subject)
         (background dir-obj))))))
```

Output:

```
[[CAT=NOUN][ROOT=kadIn][AGR=3SG][POSS=NONE][CASE=DAT]] --> kadIna
[[CAT=NOUN][ROOT=adam][AGR=3SG][POSS=NONE][CASE=NOM]] --> adam
[[CAT=VERB][ROOT=ver][SENSE=POS][TAM1=PAST][AGR=3SG]] --> verdi
[[CAT=NOUN][ROOT=elma][AGR=3SG][POSS=NONE][CASE=ACC]] --> elmayI
[PERIOD] -->.
```

**Example 3:**

Input:

```
; Adam kitabI okumak istedi.
; The direct object is a sentential clause representing an
; indefinite act.

((s-form finite)
 (clause-type predicative)
 (voice active)
 (speech-act declarative)
 (verb
  ((root "iste")
   (sense positive)
   (tense past)
   (aspect perfect)))
 (arguments
  ((subject
     ((referent
        ((arg
           ((concept "adam")))
         (agr
           ((number singular)
            (person 3))))))))
   (dir-obj
    ((roles
       ((role act)
        (arg
          ((s-form inf-ind-act)
           (clause-type predicative)
           (voice active)
           (speech-act declarative)
           (verb
             ((root "oku")
```

```
                   (sense positive)))
             (arguments
              ((dir-obj
                 ((referent
                    ((arg
                       ((concept "kitap")))
                     (agr
                       ((number singular)
                        (person 3)))))
                  (specifier
                   ((quan
                     ((definite +)))))))))))))))
```

Output:

```
[[CAT=NOUN][ROOT=adam][AGR=3SG][POSS=NONE][CASE=NOM]] --> adam
[[CAT=NOUN][ROOT=kitap][AGR=3SG][POSS=NONE][CASE=ACC]] --> kitabI
[[CAT=VERB][ROOT=oku][SENSE=POS][CONV=NOUN=MAK][TYPE=INFINITIVE]
  [AGR=3SG][POSS=NONE][CASE=NOM]] --> okumak
[[CAT=VERB][ROOT=iste][SENSE=POS][TAM1=PAST][AGR=3SG]] --> istedi
[PERIOD] --> .
```

**Example 4:**

Input:

```
; Adam kadInIn kitap okuduGunu zannetti.
; The direct object is a sentential clause representing a fact.

((s-form finite)
 (clause-type predicative)
 (voice active)
 (speech-act declarative)
 (verb
  ((root "zanned")
   (sense positive)
   (tense past)
   (aspect perfect)))
 (arguments
  ((subject
     ((referent
        ((arg
           ((concept "adam")))
         (agr
           ((number singular)
            (person 3))))))
    (dir-obj
     ((roles
        ((role fact)
         (arg
           ((s-form participle)
            (clause-type predicative)
            (voice active)
            (speech-act declarative)
            (verb
              ((root "oku")
               (sense positive)
```

```
                    (tense past)))
              (arguments
               ((dir-obj
                  ((referent
                     ((arg
                        ((concept "kitap")))
                      (agr
                        ((number singular)
                         (person 3)))))
                   (specifier
                    ((quan
                       ((definite -)))))))
                (subject
                 ((referent
                    ((arg
                       ((concept "kadIn")))
                     (agr ((number singular)(person 3)))))))))))
```

Output:

```
[[CAT=NOUN][ROOT=adam][AGR=3SG][POSS=NONE][CASE=NOM]] --> adam
[[CAT=NOUN][ROOT=kadIn][AGR=3SG][POSS=NONE][CASE=GEN]] --> kadInIn
[[CAT=NOUN][ROOT=kitap][AGR=3SG][POSS=NONE][CASE=NOM]] --> kitap
[[CAT=VERB][ROOT=oku][SENSE=POS][CONV=NOUN=DIK]
 [AGR=3SG][POSS=3SG][CASE=ACC]] --> okuduGunu
[[CAT=VERB][ROOT=zanned][SENSE=POS][TAM1=PAST][AGR=3SG]] --> zannetti
[PERIOD] --> .
```

**Example 5:**

Input:

```
; Adam kitabI okuyan kadIna elma verdi.
; The goal has a sentential modifier, which is a gapped sentence.
; The gap of the sentential clause is its theme.

((s-form finite)
 (clause-type predicative)
 (voice active)
 (speech-act declarative)
 (verb
  ((root "ver")
   (sense positive)
   (tense past)
   (aspect perfect)))
 (arguments
  ((subject
     ((referent
        ((arg
           ((concept "adam")))
         (agr
           ((number singular)
            (person 3))))))))
   (goal
     ((roles
        ((role agent)
         (arg
           ((s-form participle)
            (clause-type predicative)
            (voice active)
            (speech-act declarative)
            (verb
              ((root "oku")
```

```
                    (sense positive)
                    (tense past)))
              (arguments
               ((dir-obj
                  ((referent
                     ((arg
                        ((concept "kitap")))
                      (agr
                        ((number singular)
                         (person 3)))))
                   (specifier
                    ((quan
                       ((definite +)))))))
                (subject
                 ((referent
                    ((arg
                       ((concept" kadIn")))
                     (agr
                       ((number singular)
                        (person 3)))))))))))))
        (dir-obj
         ((referent
            ((arg
               ((concept "elma")))
             (agr
               ((number singular)
                (person 3)))))
          (specifier
           ((quan
              ((definite -)))))))))

Output:

  [[CAT=NOUN][ROOT=adam][AGR=3SG][POSS=NONE][CASE=NOM]] --> adam
  [[CAT=NOUN][ROOT=kitap][AGR=3SG][POSS=NONE][CASE=ACC]] --> kitabI
```

```
[[CAT=VERB][ROOT=oku][SENSE=POS][CONV=ADJ=YAN]] --> okuyan
[[CAT=NOUN][ROOT=kadIn][AGR=3SG][POSS=NONE][CASE=DAT]] --> kadIna
[[CAT=NOUN][ROOT=elma][AGR=3SG][POSS=NONE][CASE=NOM]] --> elma
[[CAT=VERB][ROOT=ver][SENSE=POS][TAM1=PAST][AGR=3SG]] --> verdi
[PERIOD] -->.
```

**Example 6:**

Input:

```
; Adam kadInIn okuduGu kitabI istedi.
; The direct object has a sentential modifier.

((s-form finite)
 (clause-type predicative)
 (voice active)
 (speech-act declarative)
 (verb
  ((root "iste")
   (sense positive)
   (tense past)
   (aspect perfect)))
 (arguments
  ((subject
     ((referent
        ((arg
           ((concept "adam")))
         (agr
          ((number singular)
           (person 3)))))))
   (dir-obj
    ((roles
       ((role theme)
        (arg
          ((s-form participle)
           (clause-type predicative)
           (voice active)
           (speech-act declarative)
           (verb
             ((root "oku")
              (sense positive)
```

```
                           (tense past)))
                  (arguments
                   ((dir-obj
                      ((referent
                         ((arg
                            ((concept "kitap")))
                          (agr
                            ((number singular)
                             (person 3)))))
                        (specifier
                         ((quan
                            ((definite +)))))))
                    (subject
                     ((referent
                         ((arg
                            ((concept "kadIn")))
                          (agr
                            ((number singular)
                             (person 3))))))))))))))))
```

Output:

```
[[CAT=NOUN][ROOT=adam][AGR=3SG][POSS=NONE][CASE=NOM]] --> adam
[[CAT=NOUN][ROOT=kadIn][AGR=3SG][POSS=NONE][CASE=GEN]] --> kadInIn
[[CAT=VERB][ROOT=oku][SENSE=POS][CONV=ADJ=DIK][POSS=3SG]] --> okuduGu
[[CAT=NOUN][ROOT=kitap][AGR=3SG][POSS=NONE][CASE=ACC]] --> kitabI
[[CAT=VERB][ROOT=iste][SENSE=POS][TAM1=PAST][AGR=3SG]] --> istedi
[PERIOD] --> .
```

**Example 7:**

Input:

```
; Adam eve gelir gelmez yattI.
; Expression of time is a sentential clause representing an
; adverbial.

((s-form finite)
 (clause-type predicative)
 (voice active)
 (speech-act declarative)
 (verb
  ((root "yat")
   (sense positive)
   (tense past)
   (aspect perfect)))
 (arguments
  ((subject
     ((referent
        ((arg
           ((concept "adam")))
         (agr
           ((number singular)
            (person 3))))))))
   (adjuncts
    ((time
       ((adv-type as-soon-as)
        (argument
          ((s-form adverbial)
           (clause-type predicative)
           (voice active)
           (speech-act declarative)
           (verb
             ((root "gel")))
```

```
            (arguments
             ((goal
                ((referent
                   ((arg
                      ((concept "ev")))
                    (agr
                      ((number singular)
                       (person 3)))))))))))))
```

Output:

```
[[CAT=NOUN][ROOT=adam][AGR=3SG][POSS=NONE][CASE=NOM]] --> adam
[[CAT=NOUN][ROOT=ev][AGR=3SG][POSS=NONE][CASE=DAT]] --> eve
[[CAT=VERB][ROOT=gel][SENSE=POS][TAM1=AORIST][AGR=3SG]] --> gelir
[[CAT=VERB][ROOT=gel][SENSE=NEG][TAM1=AORIST][AGR=3SG]] --> gelmez
[[CAT=VERB][ROOT=yat][SENSE=POS][TAM1=PAST][AGR=3SG]] --> yattI
[PERIOD] --> .
```

**Example 8:**

Input:

```
; Adam kadInIn kitabI okumak istediGini zannetti.
; The direct object is a sentential clause representing a definite
; act, and the direct object of the sentential clause is also a
; sentential clause representing an indefinite act.

((s-form finite)
 (clause-type predicative)
 (voice active)
 (speech-act declarative)
 (verb
  ((root "zanned")
   (sense positive)
   (tense past)
   (aspect perfect)))
 (arguments
  ((subject
     ((referent
        ((arg
           ((concept "adam")))
         (agr
           ((number singular)
            (person 3)))))))
   (dir-obj
     ((roles
        ((role fact)
         (arg
           ((s-form participle)
            (clause-type predicative)
            (voice active)
            (speech-act declarative)
            (verb
```

```
          ((root "iste")
           (sense positive)
           (tense past)
           (aspect perfect)))
         (arguments
          ((dir-obj
             ((roles
                ((role act)
                 (arg
                   ((s-form inf-ind-act)
                    (clause-type predicative)
                    (voice active)
                    (speech-act declarative)
                    (verb
                      ((root "oku")
                       (sense positive)))
                    (arguments
                      ((dir-obj
                         ((referent
                            ((arg
                               ((concept "kitap")))
                             (agr
                               ((number singular)
                                (person 3)))))
                          (specifier
                            ((quan
                               (definite +)))))))))))))
           (subject
             ((referent
                ((arg
                   ((concept "kadIn")))
                 (agr
                   ((number singular)
                    (person 3))))))))))))))))
```

Output:

```
[[CAT=NOUN][ROOT=adam][AGR=3SG][POSS=NONE][CASE=NOM]] --> adam
[[CAT=NOUN][ROOT=kadIn][AGR=3SG][POSS=NONE][CASE=GEN]] --> kadInIn
[[CAT=NOUN][ROOT=kitap][AGR=3SG][POSS=NONE][CASE=ACC]] --> kitabI
[[CAT=VERB][ROOT=oku][SENSE=POS][CONV=NOUN=MAK][TYPE=INFINITIVE]
 [AGR=3SG][POSS=NONE][CASE=NOM]] --> okumak
[[CAT=VERB][ROOT=iste][SENSE=POS][CONV=NOUN=DIK]
 [AGR=3SG][POSS=3SG][CASE=ACC]] --> istediGini
[[CAT=VERB][ROOT=zanned][SENSE=POS][TAM1=PAST][AGR=3SG]] --> zannetti
[PERIOD] --> .
```

**Example 9:**

```
Input:

  ; Kitap okuyan adam kadInIn susmasInI istedi
  ; The subject has a sentential modifier, the direct object is a
  ; sentential clause representing a definite act.

 ((s-form finite)
  (clause-type predicative)
  (voice active)
  (speech-act declarative)
  (verb
   ((root "iste")
    (sense positive)
    (tense past)
    (aspect perfect)))
  (arguments
   ((subject
      ((referent
         ((agr
            ((number singular)
             (person 3)))))
       (roles
        ((role agent)
         (arg
           ((s-form participle)
            (clause-type predicative)
            (voice active)
            (speech-act declarative)
            (verb
             ((root "oku")
              (sense positive)
              (tense past)))
```

```
            (arguments
             ((dir-obj
                ((referent
                   ((arg
                      ((concept "kitap")))
                    (agr
                      ((number singular)
                       (person 3)))))
                 (specifier
                   ((quan
                      ((definite -))))))))
              (subject
                ((referent
                   ((arg
                      ((concept "adam")))
                    (agr
                      ((number singular)
                       (person 3)))))))))))))))
       (dir-obj
        ((roles
           ((role act)
            (arg
              (s-form inf-def-act)
              (clause-type predicative)
              (voice active)
              (speech-act declarative)
              (verb
                ((root "sus")
                 (sense positive)))
              (arguments
                ((subject
                   ((referent
                      ((arg
                         ((concept "kadIn")))
                       (agr
```

```
                   ((number singular)
                    (person 3)))))))))))))))))
```

Output:

```
[[CAT=NOUN][ROOT=kitap][AGR=3SG][POSS=NONE][CASE=NOM]] --> kitap
[[CAT=VERB][ROOT=oku][SENSE=POS][CONV=ADJ=YAN]] --> okuyan
[[CAT=NOUN][ROOT=adam][AGR=3SG][POSS=NONE][CASE=NOM]] --> adam
[[CAT=NOUN][ROOT=kadIn][AGR=3SG][POSS=NONE][CASE=GEN]] --> kadInIn
[[CAT=VERB][ROOT=sus][SENSE=POS][CONV=NOUN=MA][TYPE=INFINITIVE]
 [AGR=3SG][POSS=3SG][CASE=ACC]] --> susmasInI
[[CAT=VERB][ROOT=iste][SENSE=POS][TAM1=PAST][AGR=3SG]] --> istedi
[PERIOD] --> .
```

## 5.2  Future Work

Currently our grammar can not handle discontinuous constituents and certain long distance scramblings. Although these are seen very rarely in Turkish text, our grammar can be extended to handle these. The information can also be planned before being sent to the generator. This planning can be done using syntactic cues in the source language in machine translation [7, 19] or using Centering Theory [20] and given versus new information [11].

We have designed and implemented our generator taking into consideration that, it can also be used for interlingual machine translation or with a strategic generator, as demonstrated in Figure 5.1.

**Interlingual MT:**

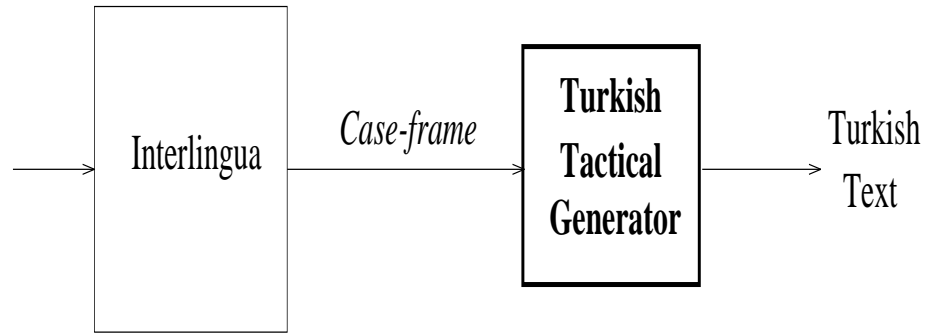

**Other Generation Applications:**

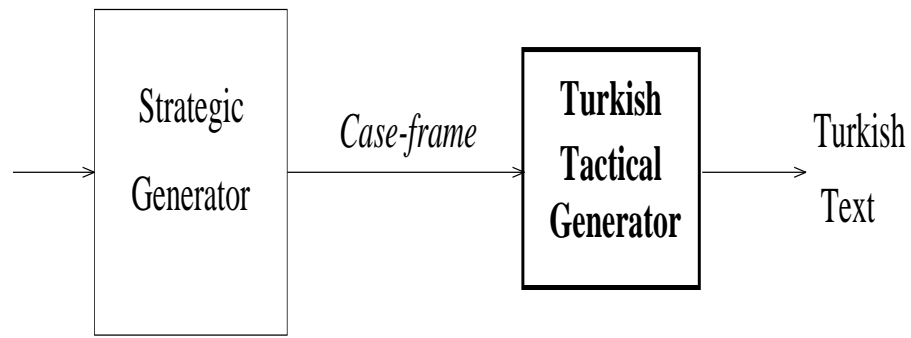

Figure 5.1: Future work.

# Chapter 6

# Conclusions

In this thesis, we have presented the highlights of our work on tactical generation in Turkish – a free constituent order language with agglutinative word structures. In addition to the content information, our generator takes as input the information structure of the sentence (topic, focus and background) and uses these to select the appropriate word order. In the absence of any information structure constituents, our generator generates sentences in a default order.

Our grammar uses a right-linear rule backbone which implements a (recursive) finite state machine for dealing with alternative word orders. The recursive behaviour of this finite state machine comes from the fact that, individual adjunct or argument constituents can also embed sentential clauses. These sentential clauses are generated using the same transitions with the sentences.

We have also provided for constituent order and stylistic variations within noun phrases based on certain emphasis and formality features. Our aim was to build a wide-coverage generator for Turkish for use in a machine translation application. We plan to use this generator in a prototype transfer-based human assisted machine translation system from English to Turkish, where the domain of translation is computer manuals.

We have designed the tactical generator taking into consideration that it can also be used in interlingual machine translation or with a strategic generator. Since concepts like long distance scramblings and discontinuous constituents are rarely seen in technical documents, we opted not to deal with them, but our grammar can be extended to cover these.

# Appendix A

# A List of Suffixes Making Adverbials

The following table lists the suffixes making temporal and manner adverbials:

| Suffix | Roles | Example |
| --- | --- | --- |
| +r ...+mAz | Temporal | Adam eve gelir gelmez uyudu. |
| +dH ...+AlH | Temporal, Manner | Annem gitti gideli uyuyamıyorum. |
| +dHkçA | Temporal, Manner | Küçük kırmızı top gittikçe hızlandı. |
| +ken | Temporal, Manner | Kitap okurken uyumuşum. |
| +mAdAn | Temporal, Manner | Sen uyumadan gideyim. |
| +yHp | Temporal, Manner | Kızınca yürüyüp gitti. |
| +ArAk | Manner | Koşarak uzaklaştı. |
| +cAsInA | Manner | Uçrcasına dışarı çıktı. |
| +A ...+A | Manner, Temporal | Koşa koşa gitti. |
| +yIncA | Manner, Temporal | Ben gelince o gitti. |
| +yAlI | Temporal | Sen geleli o uyuyor. |

# Appendix B

# Gapped Sentential Clauses

The following tables demonstrate the verb forms of sentential clauses in different situations:

| Role | Voice | Tense | Trans. | Subject | Part-Form | Example |
|---|---|---|---|---|---|---|
| Agent Exper. | Act. | Past/ Pres. | Trans. | spec./non-spec. | +yAn | Çocuğa elma satan adam |
| | | | Intrans. | - | +yAn | Çok uyuyan çocuk |
| | | Fut. | Trans. | spec./non-spec. | +yAcAk | Çocuğa elmayı satacak adam |
| | | | Intrans. | - | +yAcAk | Uyuyacak çocuk |
| | | Past+ Narr. | Trans. | spec./non-spec. | +mHş | Yemeğini yemiş çocuk |
| | | | Intrans. | - | +mHş | Okumuş çocuk |
| | Pass. | Past/ Pres. | Trans. | spec./non-spec. | – | – |
| | | | Intrans. | - | – | – |
| | | Fut. | Trans. | spec./non-spec. | – | – |
| | | | Intrans. | - | – | – |
| | | Past+ Narr. | Trans. | spec./non-spec. | – | – |
| | | | Intrans. | - | – | – |
| | Caus. | Past/ Pres. | Trans. | spec./non-spec. | +yAn | Çocuğa yemek yediren kadın |
| | | | Intrans. | - | +yAn | Çocuğu uyutan kadın |
| | | Fut. | Trans. | spec./non-spec. | +yAcAk | Çocuğa yemek yedirecek kadın |
| | | | Intrans. | - | +yAcAk | Çocuğu uyutacak kadın |
| | | Past+ Narr. | Trans. | spec./non-spec. | +mHş | Çocuğa yemek yedirmiş kadın |
| | | | Intrans. | - | +mHş | Çocuğu uyutmuş kadın |

| Role | Voice | Tense | Trans. | Subject | Part-Form | Example |
|---|---|---|---|---|---|---|
| Patient Theme | Act. | Past/ Pres. | Trans. | spec./non-spec. | +dHk+POSS | Adamın okuduğu kitap |
| | | | Intrans. | - | – | – |
| | | Fut. | Trans. | spec./non-spec. | +yAcAk+POSS | Adamın okuyacağı kitap |
| | | | Intrans. | - | – | – |
| | Pass. | Past/ Pres. | Trans. | spec./non-spec. | +yAn | Adam tarafından okunan kitap |
| | | | Intrans. | - | – | – |
| | | Fut. | Trans. | spec./non-spec. | +yAcAk | Adam tarafından okunacak kitap |
| | | | Intrans. | - | – | – |
| | Caus. | Past/ Pres. | Trans. | spec./non-spec. | +dHk+POSS | Kadının çocuğa yedirdiği yemek |
| | | | Intrans. | - | – | – |
| | | Fut. | Trans. | spec./non-spec. | +yAcAk+POSS | Kadının çocuğa yedireceği yemek |
| | | | Intrans. | - | – | – |
| Source | Active | Past/ Pres. | Trans. | spec./non-spec. | +dHk+POSS | Çocuğun denize taş attığı iskele |
| | | | Intrans. | - | +dHk+POSS | Çocuğun eve yürüdüğü okul |
| | | Fut. | Trans. | spec./non-spec. | +yAcAk+POSS | Çocuğun denize taş atacağı iskele |
| | | | Intrans. | - | +yAcAk+POSS | Çocuğun eve yürüyeceği okul |
| | Passive | Past/ Pres. | Trans. | spec. | +dHk+POSS | Taşın denize atıldığı iskele |
| | | | | non-spec. | +yAn | Denize taş atılan iskele |
| | | | Intrans. | - | +yAn | Okula yürünen ev |
| | | Fut. | Trans. | spec. | +yAcAk+POSS | Taşın denize atılacağı iskele |
| | | | | non-spec. | +yAcAk | Denize taş atılacak iskele |
| | | | Intrans. | - | +yAcAk | Okula yürünecek ev |
| | Caus. | Past/ Pres. | Trans. | spec./non-spec. | +dHk+POSS | Kadının denize taş attırdığı iskele |
| | | | Intrans. | - | +dHk+POSS | Kadının çocuğu okula yürüttüğü ev |
| | | Fut. | Trans. | spec./non-spec. | +yAcAk+POSS | Kadının denize taş attıracağı iskele |
| | | | Intrans. | - | +yAcAk+POSS | Kadının çocuğu okula yürüteceği ev |
| Goal | Act. | Past/ Pres. | Trans. | spec./non-spec. | +dHk+POSS | Çocuğun iskeleden taş attığı deniz |
| | | | Intrans. | - | +dHk+POSS | Çocuğun okuldan yürüdüğü ev |
| | | Fut. | Trans. | spec./non-spec. | +yAcAk+POSS | Çocuğun iskeleden taş atacağı deniz |
| | | | Intrans. | - | +yAcAk+POSS | Çocuğun okuldan yürüyeceği ev |
| | Pass. | Past/ Pres. | Trans. | spec. | +dHk+POSS | Taşın iskeleden atıldığı deniz |
| | | | | non-spec. | +yAn | İskeleden taş atılan deniz |
| | | | Intrans. | - | +yAn | Evden yürünen okul |
| | | Fut. | Trans. | spec. | +yAcAk+POSS | Taşın iskeleden atılacağı deniz |
| | | | | non-spec. | +yAcAk | İskeleden taş atılacak deniz |
| | | | Intrans. | - | +yAcAk | Evden yürünecek okul |
| | Caus. | Past/ Pres. | Trans. | spec./non-spec. | +dHk+POSS | Kadının iskeleden taş attırdığı deniz |
| | | | Intrans. | - | +dHk+POSS | Kadının çocuğu evden yürüttüğü okul |
| | | Fut. | Trans. | spec./non-spec. | +yAcAk+POSS | Kadının iskeleden taş attıracağı deniz |
| | | | Intrans. | - | +yAcAk+POSS | Kadının çocuğu yürüteceği okul |

| Role | Voice | Tense | Trans. | Subject | Part-Form | Example |
|---|---|---|---|---|---|---|
| Loc. | Active | Past/ Pres. | Trans. | spec./non-spec. | +dHk+POSS | Çocuğun kitap okuduğu masa |
| | | | Intrans. | - | +dHk+POSS | Çocuğun uyuduğu yatak |
| | | Fut. | Trans. | spec./non-spec. | +yAcAk+POSS | Çocuğun kitap okuyacağı masa |
| | | | Intrans. | - | +yAcAk+POSS | Çocuğun okuldan uyuyacağı yatak |
| | Passive | Past/ Pres. | Trans. | spec. | +dHk+POSS | kitabın okunduğu masa |
| | | | | non-spec. | +yAn | kitap okunan masa |
| | | | Intrans. | - | +yAn | uyunan yatak |
| | | Fut. | Trans. | spec. | +yAcAk+POSS | kitabın okunacağı masa |
| | | | | non-spec. | +yAcAk | kitap okunacak masa |
| | | | Intrans. | - | +yAcAk | Uyunacak yatak |
| | Caus. | Past/ Pres. | Trans. | spec./non-spec. | +dHk+POSS | Kadının çocuğa yemek yedirdiği masa |
| | | | Intrans. | - | +dHk+POSS | Kadının çocuğu evden uyuttuğu masa |
| | | Fut. | Trans. | spec./non-spec. | +yAcAk+POSS | Kadının çocuğa yemek yedireceği masa |
| | | | Intrans. | - | +yAcAk+POSS | Kadının çocuğu uyutacağı yatak |
| Benef. | Active | Past/ Pres. | Trans. | spec./non-spec. | +dHk+POSS | Kadının kitap okuduğu çocuk |
| | | | Intrans. | - | – | – |
| | | Fut. | Trans. | spec./non-spec. | +yAcAk+POSS | Kadının kitap okuyacağı çocuk |
| | | | Intrans. | - | – | – |
| | Passive | Past/ Pres. | Trans. | spec. | +dHk+POSS | Kitabın okunduğu çocuk |
| | | | | non-spec. | +yAn | Kitap okunan çocuk |
| | | | Intrans. | - | – | – |
| | | Fut. | Trans. | spec. | +yAcAk+POSS | Kitabın okunacağı çocuk |
| | | | | non-spec. | +yAcAk | Kitap okunacak çocuk |
| | | | Intrans. | - | – | – |
| | Caus. | Past/ Pres. | Trans. | spec./non-spec. | – | – |
| | | | Intrans. | - | – | – |
| | | Fut. | Trans. | spec./non-spec. | – | – |
| | | | Intrans. | - | – | – |
| C-obj | Caus. | Past/ Pres. | Trans. | spec./non-spec. | +dHk+POSS | Kadının yemek yedirdiği çocuk |
| | | | Intrans. | - | +dHk+POSS | Kadının uyuttuğu çocuk |
| | | Fut. | Trans. | spec./non-spec. | +yAcAk+POSS | Kadının yemek yedireceği çocuk |
| | | | Intrans. | - | +yAcAk+POSS | Kadının uyutacağı çocuk |

| Role | Voice | Tense | Trans. | Subject | Part-Form | Example |
|---|---|---|---|---|---|---|
| Rec. | Active | Past/Pres. | Trans. | spec./non-spec. | +dHk+POSS | Kadının kitap verdiği çocuk |
| | | | Intrans. | - | – | – |
| | | Fut. | Trans. | spec./non-spec. | +yAcAk+POSS | Kadının kitap vereceği çocuk |
| | | | Intrans. | - | – | – |
| | Passive | Past/Pres. | Trans. | spec. | +dHk+POSS | Kitabın verildiği çocuk |
| | | | | non-spec. | +yAn | Kitap verilen çocuk |
| | | | Intrans. | - | – | – |
| | | Fut. | Trans. | spec. | +yAcAk+POSS | Kitabın verileceği çocuk |
| | | | | non-spec. | +yAcAk | Kitap verilecek çocuk |
| | | | Intrans. | - | – | – |
| | Caus. | Past/Pres. | Trans. | spec./non-spec. | +dHk+POSS | Kadının adama kitap verdirdiği çocuk |
| | | | Intrans. | - | – | – |
| | | Fut. | Trans. | spec./non-spec. | +yAcAk+POSS | Kadının adama kitap verdireceği çocuk |
| | | | Intrans. | - | – | – |
| Time *Dur.* | Active | Past/Pres. | Trans. | spec./non-spec. | +dHk+POSS | Adamın elmayı sattığı zaman/*süre* |
| | | | Intrans. | - | +dHk+POSS | Çocuğun uyuduğu zaman/*süre* |
| | | Fut. | Trans. | spec./non-spec. | +yAcAk+POSS | Adamın elmayı satacağı zaman/*süre* |
| | | | Intrans. | - | +yAcAk+POSS | Çocuğun uyuyacağı zaman/*süre* |
| | Passive | Past/Pres. | Trans. | spec. | +dHk+POSS | Kitabın okunduğu zaman/*süre* |
| | | | | non-spec. | +yAn | Kitap okunan zaman/*süre* |
| | | | Intrans. | - | +dHk+POSS | Okula yüründüğü zaman |
| | | Fut. | Trans. | spec. | +yAcAk+POSS | Kitabın okunacağı zaman/*süre* |
| | | | | non-spec. | +yAcAk | Kitap okunacak zaman/*süre* |
| | | | Intrans. | - | +yAcAk | Okula yürünecek zaman |
| | Caus. | Past/Pres. | Trans. | spec./non-spec. | +dHk+POSS | Kadının yemek yedirdiği zaman/*süre* |
| | | | Intrans. | - | +dHk+POSS | Kadının çocuğu uyuttuğu zaman/*süre* |
| | | Fut. | Trans. | spec./non-spec. | +yAcAk+POSS | Kadının yemek yedireceği zaman/*süre* |
| | | | Intrans. | - | +yAcAk+POSS | Kadının çocuğu uyutacağı zaman/*süre* |
| Place | Active | Past/Pres. | Trans. | spec./non-spec. | +dHk+POSS | Çocuğun yemek yediği masa |
| | | | Intrans. | - | +dHk+POSS | Çocuğun yürüdüğü yol |
| | | Fut. | Trans. | spec./non-spec. | +yAcAk+POSS | Çocuğun yemek yiyeceği masa |
| | | | Intrans. | - | +yAcAk+POSS | Çocuğun yürüyeceği yol |
| | Passive | Past/Pres. | Trans. | spec. | +dHk+POSS | Yemeğin yendiği masa |
| | | | | non-spec. | +yAn | Yemek yenilen masa |
| | | | Intrans. | - | +yAn | Yürünen yol |
| | | Fut. | Trans. | spec. | +yAcAk+POSS | Yemeğin yeneceği masa |
| | | | | non-spec. | +yAcAk | Yemek yenecek masa |
| | | | Intrans. | - | +yAcAk | Yürünecek yol |
| | Caus. | Past/Pres. | Trans. | spec./non-spec. | +dHk+POSS | Kadının çocuğa yemek yedirdiği masa |
| | | | Intrans. | - | +dHk+POSS | Kadının çocuğu yürüttüğü yol |
| | | Fut. | Trans. | spec./non-spec. | +yAcAk+POSS | Kadının çocuğa yemek yedireceği masa |
| | | | Intrans. | - | +yAcAk+POSS | Kadının çocuğu yürüteceği yol |

# Appendix C

# Examples from Computer Manuals

**Example 1:**

Input:

```
; Dosyalardaki belgelerle CalISma gOrUntUsU her gOsterildiGinde,
; dosya bilgi isteminde, en son kullandIGInIz dosyanIn adI yer alIr

((s-form finite)
 (clause-type predicative)
 (voice active)
 (speech-act declarative)
 (verb
  ((root "yer-al")
   (sense positive)
   (tense present)
   (aspect aorist)))
 (arguments
  ((subject
     ((referent
        ((arg
           ((concept "ad")))
         (agr
           ((number singular)
            (person 3)))))
```

```
        (specifier
         ((quan
           ((definite +)))))
        (possessor
         ((roles
           ((role theme)
            (arg
             ((s-form participle)
              (clause-type predicative)
              (voice active)
              (speech-act declarative)
              (verb
               ((root "kullan")
                (sense positive)
                (tense past)))
              (arguments
               ((subject
                 ((referent
                   ((agr
                     ((number plural)
                      (person 2)))))))
                (dir-obj
                 ((referent
                   ((arg
                     ((concept "dosya")))
                    (agr
                     ((number singular)
                      (person 3)))))
                  (specifier
                   ((quan
                     ((definite +)))))))))
              (adjuncts
               ((time
                 ((p-name "son")
                  (intensifier
                   ((degree "en"))))))))))))))
   (control
    ((is
      ((focus subject)))))
   (adjuncts
    ((place
      ((referent
```

```
             ((arg
               ((referent
                  ((arg
                     ((concept "istem")))
                    (agr
                     ((number singular)
                      (person 3)))))
                 (classifier
                  ((referent
                     ((arg
                        ((concept "bilgi")))
                       (agr
                        ((number singular)
                         (person 3)))))))))
                (agr
                 ((number singular)
                  (person 3)))))
           (classifier
             ((referent
               ((arg
                  ((concept "dosya")))
                 (agr
                  ((number singular)
                   (person 3)))))))))
       (time
        ((roles
           ((role fact)
            (arg
              ((s-form participle)
               (clause-type predicative)
               (voice passive)
               (speech-act declarative)
               (verb
                ((root "gOster")
                 (sense positive)
                 (tense past)))
               (arguments
                ((subject
                   ((referent
                      ((arg
                         ((concept "gOrUntU")))
                        (agr
```

```
                       ((number singular)
                         (person 3)))))
               (specifier
                ((quan
                   ((definite -)))))
               (classifier
                ((referent
                   ((roles
                      ((role act)
                       (arg
                         ((s-form inf-def-act)
                          (clause-type predicative)
                          (voice active)
                          (speech-act declarative)
                          (verb
                            ((root "CalIS")
                             (sense positive)))
                          (arguments
                            ((instrument
                               ((referent
                                  ((arg
                                     ((concept "belge")))
                                   (agr
                                     ((number plural)
                                      (person 3)))))
                                (specifier
                                  ((spec-rel
                                     ((relation location)
                                      (argument
                                        ((referent
                                           ((arg
                                              ((concept "dosya")))
                                            (agr
                                              ((number plural)
                                               (person 3)))
                                            (sem ((temporal -)))))))))))))))))
                                                       ))))))))))))
         (adjuncts
          ((manner
             ((quantifier
                ((root her))))))))))))))))))
```

```
Output:

  [[CAT=NOUN][ROOT=dosya][AGR=3PL][POSS=NONE][CASE=LOC]
   [CONV=ADJ=REL]] --> dosyalardaki
  [[CAT=NOUN][ROOT=belge][AGR=3PL][POSS=NONE][CASE=INS]] --> belgelerle
  [[CAT=VERB][ROOT=CalIS][SENSE=POS][CONV=NOUN=MA][TYPE=INFINITIVE]
   [AGR=3SG][POSS=NONE][CASE=NOM]] --> CalISma
  [[CAT=NOUN][ROOT=gOrUntU][AGR=3SG][POSS=3SG][CASE=NOM]] --> gOrUntUsU
  [[CAT=ADJ][ROOT=her][TYPE=DETERMINER]] --> her
  [[CAT=VERB][ROOT=gOster][VOICE=PASS][SENSE=POS][CONV=NOUN=DIK]
   [AGR=3SG][POSS=3SG][CASE=LOC]] --> gOsterildiGinde
  [[CAT=NOUN][ROOT=dosya][AGR=3SG][POSS=NONE][CASE=NOM]] --> dosya
  [[CAT=NOUN][ROOT=bilgi][AGR=3SG][POSS=NONE][CASE=NOM]] --> bilgi
  [[CAT=NOUN][ROOT=istem][AGR=3SG][POSS=3SG][CASE=LOC]] --> isteminde
  [[CAT=ADVERB][ROOT=en]] --> en
  [[CAT=ADJ][ROOT=son]] --> son
  [[CAT=VERB][ROOT=kullan][SENSE=POS][CONV=ADJ=DIK][POSS=2PL]]
  --> kullandIGInIz
  [[CAT=NOUN][ROOT=dosya][AGR=3SG][POSS=NONE][CASE=GEN]] --> dosyanIn
  [[CAT=NOUN][ROOT=ad][AGR=3SG][POSS=3SG][CASE=NOM]] --> adI
  [[CAT=VERB][ROOT=yer-al][SENSE=POS][TAM1=AORIST][AGR=3SG]] --> yer-alIr
  [PERIOD] --> .
```

**Example 2:**

```
Input:

  ; kullanIm tanItIm alanInda belge biCimini tanImlayan metin
  ; tanItImInIn adI yer alIr

  ((s-form finite)
   (clause-type predicative)
   (voice active)
   (speech-act declarative)
   (verb
    ((root "yer-al")
     (sense positive)
     (tense present)
     (aspect aorist)))
   (control
```

```
          ((is
            ((focus subject)))))
         (adjuncts
          ((place
            ((referent
               ((arg
                  ((referent
                     ((arg
                        ((concept "alan")))
                      (agr
                        ((number singular)
                         (person 3)))))
                    (classifier
                     ((referent
                        ((arg
                           ((concept "tanItIm")))
                         (agr
                           ((number singular)
                            (person 3)))))))))
                (agr
                  ((number singular)
                   (person 3)))))
              (classifier
               ((referent
                  ((arg
                     ((concept "kullanIm")))
                   (agr
                     ((number singular)
                      (person 3)))))))))))
         (arguments
          ((subject
            ((referent
               ((arg
                  ((concept "ad")))
                (agr
                  ((number singular)
                   (person 3)))))
              (possessor
               ((roles
                  ((role agent)
                   (arg
                     ((s-form participle)
```

```
                    (clause-type predicative)
                    (voice active)
                    (speech-act declarative)
                    (verb
                     ((root "tanImla")
                      (sense positive)
                      (tense past)))
                    (arguments
                     ((dir-obj
                        ((referent
                           ((arg
                              ((concept "biCim")))
                            (agr
                              ((number singular)
                               (person 3)))))
                         (classifier
                           ((referent
                              ((arg
                                 ((concept "belge")))
                               (agr
                                 ((number singular)
                                  (person 3)))))))))
                      (subject
                        ((referent
                           ((arg
                              ((concept "tanItIm")))
                            (agr
                              ((number singular)
                               (person 3)))))
                         (classifier
                           ((referent
                              ((arg
                                 ((concept "metin")))
                               (agr
                                 ((number singular)
                                  (person 3)))))))))))))

Output:

  [[CAT=NOUN][ROOT=kullanIm][AGR=3SG][POSS=NONE][CASE=NOM]] --> kullanIm
  [[CAT=NOUN][ROOT=tanItIm][AGR=3SG][POSS=NONE][CASE=NOM]] --> tanItIm
  [[CAT=NOUN][ROOT=alan][AGR=3SG][POSS=3SG][CASE=LOC]] --> alanInda
```

```
[[CAT=NOUN][ROOT=belge][AGR=3SG][POSS=NONE][CASE=NOM]] --> belge
[[CAT=NOUN][ROOT=biCim][AGR=3SG][POSS=3SG][CASE=ACC]] --> biCimini
[[CAT=VERB][ROOT=tanImla][SENSE=POS][CONV=ADJ=YAN]] --> tanImlayan
[[CAT=NOUN][ROOT=metin][AGR=3SG][POSS=NONE][CASE=NOM]] --> metin
[[CAT=NOUN][ROOT=tanItIm][AGR=3SG][POSS=3SG][CASE=GEN]] --> tanItImInIn
[[CAT=NOUN][ROOT=ad][AGR=3SG][POSS=3SG][CASE=NOM]] --> adI
[[CAT=VERB][ROOT=yer-al][SENSE=POS][TAM1=AORIST][AGR=3SG]] --> yer-alIr
[PERIOD] --> .
```

**Example 3:**

```
Input:

; Metni yazarken kullanabileceGiniz iSlev tuSlarInI iCeren
; listenin gOrUntUnUn alt bOlUmUnde gOsterilmesini seCebilirsiniz

((s-form finite)
 (clause-type predicative)
 (voice active)
 (speech-act declarative)
 (verb
  ((root "seC")
   (sense positive)
   (modality yabil)
   (tense present)
   (aspect aorist)))
 (arguments
  ((subject
     ((referent
        ((agr
           ((number plural)
            (person 2)))))))
   (dir-obj
     ((roles
        ((role act)
         (arg
           ((s-form inf-def-act)
            (clause-type predicative)
            (voice passive)
```

```
(speech-act declarative)
(verb
 ((root "gOster")
  (sense positive)))
(arguments
 ((subject
    ((referent
       ((arg
          ((concept "liste")))
         (agr
          ((number singular)
           (person 3)))))
     (roles
      ((role agent)
       (arg
         ((s-form participle)
          (clause-type predicative)
          (voice active)
          (speech-act declarative)
          (verb
           ((root "iCer")
            (sense positive)
            (tense past)))
          (arguments
           ((subject
              ((referent
                 ((arg
                    ((concept "liste")))
                   (agr
                    ((number singular)
                     (person 3)))))))
            (dir-obj
              ((roles
                 ((role theme)
                   (arg
                     ((s-form participle)
                      (clause-type predicative)
                      (voice active)
                      (speech-act declarative)
                      (verb
                       ((root "kullan")
                        (sense positive)
```

```
                        (modality yabil)
                        (tense future)))
                  (arguments
                   ((dir-obj
                      ((referent
                         ((arg
                            ((concept "tuS")))
                          (agr
                           ((number plural)
                            (person 3)))))
                       (classifier
                        ((referent
                          ((arg
                            ((concept "iSlev")))
                           (agr
                            ((number singular)
                             (person 3)))))))
                       (specifier
                        ((quan
                          ((definite +)))))))
                    (subject
                      ((referent
                         ((agr
                           ((number plural)
                            (person 2))))))))
                  (adjuncts
                   ((time
                      ((adv-type while)
                       (argument
                        ((s-form adverbial)
                         (clause-type predicative)
                         (voice active)
                         (speech-act declarative)
                         (verb
                          ((root "yaz")
                           (sense positive)
                           (tense present)
                           (aspect aorist)))
                         (arguments
                          ((dir-obj
                             ((referent
                                ((arg
```

```
                                        ((concept "metin")))
                                       (agr
                                        ((number singular)
                                         (person 3)))))
                                    (specifier
                                     ((quan
                                       ((definite +)))))))))))))
                                                     ))))))))))))))))
               (location
                ((referent
                   ((arg
                      ((concept "bOlUm")))
                    (agr
                      ((number singular)
                       (person 3)))))
                  (modifier
                    ((qualitative
                       ((p-name "alt")))))
                  (possessor
                    ((referent
                       ((arg
                          ((concept "gOrUntU")))
                        (agr
                          ((number singular)
                           (person 3))))))))))))))))))))
```

Output:

```
[[CAT=NOUN][ROOT=metin][AGR=3SG][POSS=NONE][CASE=ACC]] --> metni
[[CAT=VERB][ROOT=yaz][SENSE=POS][TAM1=AORIST][CONV=ADVERB=KEN]]
--> yazarken
[[CAT=VERB][ROOT=kullan][SENSE=POS][COMP=YABIL][CONV=ADJ=YACAK]
 [POSS=2PL]] --> kullanabileceGiniz
[[CAT=NOUN][ROOT=iSlev][AGR=3SG][POSS=NONE][CASE=NOM]] --> iSlev
[[CAT=NOUN][ROOT=tuS][AGR=3PL][POSS=3SG][CASE=ACC]] --> tuSlarInI
[[CAT=VERB][ROOT=iCer][SENSE=POS][CONV=ADJ=YAN]] --> iCeren
[[CAT=NOUN][ROOT=liste][AGR=3SG][POSS=NONE][CASE=GEN]] --> listenin
[[CAT=NOUN][ROOT=gOrUntU][AGR=3SG][POSS=NONE][CASE=GEN]]
--> gOrUntUnUn
[[CAT=ADJ][ROOT=alt]] --> alt
[[CAT=NOUN][ROOT=bOlUm][AGR=3SG][POSS=3SG][CASE=LOC]] --> bOlUmUnde
[[CAT=VERB][ROOT=gOster][VOICE=PASS][SENSE=POS][CONV=NOUN=MA]
```

```
    [TYPE=INFINITIVE][AGR=3SG][POSS=3SG][CASE=ACC]] --> gOsterilmesini
   [[CAT=VERB][ROOT=seC][SENSE=POS][COMP=YABIL][TAM1=AORIST][AGR=2PL]]
   --> seCebilirsiniz
   [PERIOD] --> .
```

**Example4:**

```
Input:

 ; gOrUntUdeki ikinci satIr, yUrUrlUkteki kenar boSluGu ve sekme
 ; noktasI ayarlarInI gOsteren OlCek satIrIdIr

 ((s-form finite)
  (clause-type attributive)
  (rel is-a)
  (voice active)
  (speech-act declarative)
  (verb
   ((root to-be)
    (sense positive)
    (tense present)
    (aspect aorist)))
  (arguments
   ((subject
      ((referent
         ((arg
            ((concept "satIr")))
          (agr
            ((number singular)
             (person 3)))))
       (modifier
         ((ordinal
            ((order 2)))))
       (specifier
         ((spec-rel
            ((relation location)
             (argument
               ((referent
                  ((arg
                     ((concept "gOrUntU")))
                   (agr
```

```
                           ((number singular)
                             (person 3)))
                         (sem ((temporal -)))))))))))))
             (pred-property
              ((roles
                 ((role agent)
                   (arg
                    ((s-form participle)
                     (clause-type predicative)
                     (voice active)
                     (speech-act declarative)
                     (verb
                      ((root "gOster")
                       (sense positive)
                       (tense past)))
                     (arguments
                      ((subject
                         ((referent
                            ((arg
                               ((concept "satIr")))
                             (agr
                               ((number singular)
                                (person 3)))))
                          (classifier
                            ((referent
                               ((arg
                                  ((concept "OlCek")))
                                (agr
                                  ((number singular)
                                   (person 3)))))))))
                       (dir-obj
                         ((conj ve)
                          (elements
                            (*multiple*
                              ((referent
                                 ((arg
                                    ((concept "boSluk")))
                                  (agr
                                    ((number singular)
                                     (person 3)))))
                               (classifier
                                 ((referent
```

```
                  ((arg
                     ((concept "kenar")))
                   (agr
                     ((number singular)
                      (person 3)))))))
              (specifier
                ((spec-rel
                   ((relation location)
                    (argument
                      ((referent
                         ((arg
                            ((concept "yUrUrlUk")))
                          (agr
                            ((number singular)
                             (person 3)))
                          (sem ((temporal -)))))))))))))
             ((referent
               ((arg
                  ((concept "ayar")))
                (agr
                  ((number plural)
                   (person 3)))))
              (classifier
                ((referent
                   ((arg
                      ((concept "nokta")))
                    (agr
                      ((number singular)
                       (person 3)))))
                 (classifier
                   ((referent
                      ((roles
                         ((role act)
                          (arg
                            ((s-form inf-def-act)
                             (clause-type predicative)
                             (voice active)
                             (speech-act declarative)
                             (verb
                               ((root "sek")
                                (sense positive)))))))))))
              (specifier
```

```
                       ((quan
                         ((definite +)))))))))
                                                   )))))))))))))
```

Output:

```
[[CAT=NOUN][ROOT=gOrUntU][AGR=3SG][POSS=NONE][CASE=LOC]
 [CONV=ADJ=REL]] --> gOrUntUdeki
[[CAT=ADJ][ROOT=ikinci][TYPE=ORDINAL]] --> ikinci
[[CAT=NOUN][ROOT=satIr][AGR=3SG][POSS=NONE][CASE=NOM]] --> satIr
[[CAT=NOUN][ROOT=yUrUrlUk][AGR=3SG][POSS=NONE][CASE=LOC]
 [CONV=ADJ=REL]] --> yUrUrlUkteki
[[CAT=NOUN][ROOT=kenar][AGR=3SG][POSS=NONE][CASE=NOM]] --> kenar
[[CAT=NOUN][ROOT=boSluk][AGR=3SG][POSS=3SG][CASE=NOM]] --> boSluGu
[[CAT=CONN][ROOT=ve]] --> ve
[[CAT=VERB][ROOT=sek][SENSE=POS][CONV=NOUN=MA][TYPE=INFINITIVE]
 [AGR=3SG][POSS=NONE][CASE=NOM]] --> sekme
[[CAT=NOUN][ROOT=nokta][AGR=3SG][POSS=3SG][CASE=NOM]] --> noktasI
[[CAT=NOUN][ROOT=ayar][AGR=3PL][POSS=3SG][CASE=ACC]] --> ayarlarInI
[[CAT=VERB][ROOT=gOster][SENSE=POS][CONV=ADJ=YAN]] --> gOrteren
[[CAT=NOUN][ROOT=OlCek][AGR=3SG][POSS=NONE][CASE=NOM]] --> OlCek
[[CAT=NOUN][ROOT=satIr][AGR=3SG][POSS=3SG][CASE=NOM]
 [CONV=VERB=NONE][TAM2=PRES][COPULA=2][AGR=3SG]] --> satIrIdIr
[PERIOD] --> .
```